\documentclass[sigconf, screen]{acmart}
\usepackage{style/sty}

\setcopyright{rightsretained}
\acmPrice{}
\acmDOI{10.1145/3368089.3409696}
\acmYear{2020}
\copyrightyear{2020}
\acmSubmissionID{fse20main-p226-p}
\acmISBN{978-1-4503-7043-1/20/11}
\acmConference[ESEC/FSE '20]{Proceedings of the 28th ACM Joint European Software Engineering Conference and Symposium on the Foundations of Software Engineering}{November 8--13, 2020}{Virtual Event, USA}
\acmBooktitle{Proceedings of the 28th ACM Joint European Software Engineering Conference and Symposium on the Foundations of Software Engineering (ESEC/FSE '20), November 8--13, 2020, Virtual Event, USA}

\usepackage[final]{changes} 
\setdeletedmarkup{\textcolor[rgb]{0.8,0,1}{\sout{#1}}}

\newcommand{\njuaffl}[0]{%
\affiliation{%
  \institution{State Key Lab of Novel Software Technology, Nanjing University}%
  \city{Nanjing}%
  \country{China}%
}%
}

\begin{document}

\title[Debugging Confidence Errors for DNNs in the Field]{Operational Calibration: Debugging Confidence Errors for DNNs in the Field}


\author[Z. Li]{Zenan Li}%
\njuaffl
\email{lizenan@smail.nju.edu.cn}
\author[X. Ma]{Xiaoxing Ma}
\authornote{Corresponding author.}
\orcid{0000-0001-7970-1384}
\njuaffl
\email{xxm@nju.edu.cn}
\author[C. Xu]{Chang Xu}
\njuaffl
\email{changxu@nju.edu.cn}
\author[J. Xu]{Jingwei Xu}
\njuaffl
\email{jingweix@nju.edu.cn}
\author[C. Cao]{Chun Cao}
\njuaffl
\email{caochun@nju.edu.cn}
\author[J. L\"{u}]{Jian L\"{u}}
\njuaffl
\email{lj@nju.edu.cn}

\begin{abstract}
Trained DNN models are increasingly adopted as integral parts of software systems, 
but they often perform deficiently in the field. 
%
A particularly damaging problem is that DNN models often give false predictions with high confidence,
due to the unavoidable slight divergences between operation data and training data. 
To minimize the loss caused by inaccurate confidence, operational calibration, i.e.,
calibrating the confidence function of a DNN classifier against its operation domain,  
becomes a necessary debugging step in the engineering of the whole system.
 
Operational calibration is difficult considering 
the limited budget of labeling operation data and the weak interpretability of DNN models. 
We propose a Bayesian approach to operational calibration 
that gradually corrects the confidence given by the model under calibration
with a small number of labeled operation data 
deliberately selected from a larger set of unlabeled operation data. 
\added{The approach is made effective and efficient by leveraging} \deleted{Exploiting} the locality of the learned representation of the DNN model and 
modeling the calibration as Gaussian Process Regression\deleted{,
the approach achieves impressive efficacy and efficiency}.
Comprehensive experiments with various practical datasets and DNN models show that 
it significantly outperformed alternative methods, 
and in some difficult tasks it eliminated about 71\% to 97\% high-confidence~($>$$0.9$) errors with only about 10\% of 
the minimal amount of labeled operation data needed for 
practical learning techniques to barely work.

\end{abstract}

%
%

\ccsdesc[500]{Software and its engineering~Software testing and debugging}
\ccsdesc[500]{Computing methodologies~Neural networks}

\keywords{Operational Calibration, Deep Neural Networks, Gaussian Process}

\maketitle

\epigraph{\emph{To know what you know and what you do not know, that is true knowledge.}}{Confucius. 551–479 BC.}


\section{Introduction}
Deep learning (DL) has achieved human-level or even better performance in some difficult tasks, 
such as image classification and speech recognition~\cite{LeCun:2015aa, Goodfellow-et-al-2016}.
Deep Neural Network (DNN) models are increasingly adopted in high-stakes application scenarios such as medical diagnostics~\cite{obermeyer2016predicting} and self-driven cars~\cite{bojarski2016selfdriven}. 
However, it is not uncommon that DNN models perform poorly in practice~\cite{Riley2019nature}.
The interest in the quality assurance for DNN models as integral parts of software systems 
is surging in the community of software engineering~\cite{pei2017deepxplore, ma2018combinatorial, sun2018concolic, zhang2018deeproad, kim2019guiding, zhang2019surveyMLT}.



\deleted{
While machine learning is primarily concerned with how well a model learns from its training data, 
software engineering must pay more attention to how well the trained model performs in the field.
This means that the quality assurance of DNN models used as software components needs to be \emph{operational}. 
In addition, it needs to be \emph{stochastic} to embrace the intrinsic uncertainty of the models 
and \emph{effective} to best utilize the effort in collecting and labeling operation data. 
}

A particular problem of using a previously \added{well-}trained DNN model in an operation domain is that 
the model may not only make more-than-expected mistakes in its predictions,  
but also give erroneous confidence values for these predictions. 
\deleted{Arguably the} \added{The} latter issue is \deleted{more harmful} \added{particularly problematic for decision making}, 
because if the confidence values were accurate, the model would be at least partially usable by accepting only high-confidence predictions.   
It needs to be emphasized that erroneous  predictions with high confidence are especially damaging 
because users will take high stakes in them. 
For example, an over-confident benign prediction for a pathology image 
could mislead a doctor into overlooking a malignant tumor.

The problem comes from the almost inevitable divergences between 
the original data on which a model is trained and the actual data in its operation domain,
which is often called \emph{domain shift} or \emph{dataset shift}~\cite{ng2016nuts}
 in the machine learning literature.
It can be difficult and go beyond the stretch of usual machine learning tricks 
such as fine-tuning and transfer learning~\cite{pan2009survey,wang14icml}, 
because of two practical restrictions often encountered. 
First, 
the training data of a third-party DNN model are often
unavailable due to privacy and proprietary limitations~\cite{zhou2016learnware, Shokri2015CCS, konevcny2016federated}.
Second, one can only use a small number of labeled operation data 
because it could be very expensive to label the data collected in the field. 
For example, in an AI-assisted clinical medicine scenario, 
surgical biopsies may have to be involved in the labeling of radiology images.

We consider \emph{operational calibration} 
that corrects the error in the confidence provided by a DNN model 
for its prediction on each input in a given operation domain. 
It does not change the predictions themselves  
but tells when the model works well and when not.
\added{As the quantification of the intrinsic uncertainty in the predictions made by a model, 
confidence values are integral parts of the model's outputs.
So operational calibration can be viewed as a kind of \emph{debugging} activity 
that identifies and fixes errors in these parts of model outputs. 
It improves the model's quality of service in the field 
with more accurate confidence for better decision making. } 
\replaced{As a quality assurance activity, operational calibration 
shall be carried out during the deployment of a previously trained DNN model 
in a new operation domain.
One can also incorporate it into the system and excise it from time to time to adapt 
the model to the evolving data distribution in the operation domain.
}{In this sense, operational calibration is a necessary debugging step 
that should be incorporated in the engineering of the whole system. 
It fits a perviously trained DNN model to its operational domain with calibrated confidence for better decision making.  
Also, one can incorporate it into the system and excise it at runtime to continuously adapt to the evolutions in the operation domain.
} 
\deleted{We note that operational calibration is challenging because what it fixes is a function, not just a value.} 

It is natural to model operational calibration as a \added{case} \deleted{kind} of non-parametric Bayesian Inference 
and solve it with Gaussian Process Regression~\cite{rasmussen2005GPML}.  
We take the original confidence of a DNN model as the prior, 
and gradually calibrate the confidence with the evidence collected by selecting and labeling operation data. 
The key insight into effective and efficient regression comes from two observations:
First, the DNN model, although suffering from the domain shift, 
can be used as a feature extractor with which unlabeled operation data
can be nicely clustered~\cite{zhu2005semi, shu2018a}. 
In each cluster, the prediction correctness of an example is correlated with another one. 
The correlation can be effectively estimated with the distance of the two examples in the feature space.
Second, Gaussian Process is able to quantify the uncertainty after each step, 
which can be used to guide the selection of operation data to label efficiently. 

Systematic empirical evaluations showed that the approach was promising. 
It \deleted{significantly} outperformed existing calibration methods in both efficacy and efficiency
in all settings we tested. 
In some difficult tasks, it eliminated about 71\% to 97\% high-confidence errors 
with only about 10\% of the minimal amount of labeled operation data 
 needed for practical learning techniques to barely work.
\deleted{In practice, such few operation data can be collected as feedback during a system's runtime, 
and largely improve the system's reliability for its later decisions to make.} 

In summary, the contributions of this paper are: 
\begin{itemize}
%
\item Examining quality assurance for  DNN models used as software components, 
and raising the problem of operational calibration as debugging for confidence errors of DNNs in the field.
\item Proposing a  Gaussian Process-based approach to operational calibration, 
which leverages the representation learned by the DNN model under calibration 
and the locality of confidence errors in this representation. 
\item Evaluating the approach systematically. Experiments with various datasets and models confirmed the general efficacy and efficiency of our approach.
\end{itemize}

The rest of this paper is organized as follows. 
We first discuss the general need for operational quality assurance for DNNs in Section~\ref{sec:operqa}, 
and then define the the problem of operational calibration in Section~\ref{sec:problem}. 
We detail our approach to operational calibration in Section~\ref{sec:method}
and evaluate it empirically in Section~\ref{sec:evaluation}. 
We overview related work and highlight their differences from ours in Section~\ref{sec:relatedwork}
before concluding the paper with Section~\ref{sec:conclusion}.


\section{Quality assurance for DNN models used as software artifacts}
\label{sec:operqa}




Well trained DNN models can provide marvelous capabilities, 
but unfortunately their failures in applications are also very common~\cite{Riley2019nature}.
When using a trained model as an integral part of a high-stakes software system, 
it is crucial to know quantitatively \emph{how well} the model will work \added{and to adapt it to the application conditions}.
The quality assurance combining the viewpoints from software engineering and machine learning is needed, but largely missing. 
\added{In what follows, we first discuss the non-conventional requirements for such quality assurance, 
and then give an application scenario to highlight the software engineering concerns. 
}

Deep learning is intrinsically inductive~\cite{Goodfellow-et-al-2016, zhou2019abductive}.
However, conventional software engineering is mostly deductive, 
as evidenced by its fundamental principle of specification-implementation consistency. 
A specification defines the assumptions and guarantees of a software artifact.
The artifact is expected to meet its guarantees whenever its assumptions are satisfied.
Thus explicit specifications make software artifacts more or less domain independent.
However, statistical machine learning does not provide such kind of specifications. 
Essentially it tries to induce a model from its training data, 
which is intended to be general 
so that the model can give predictions on previously unseen inputs. 
Unfortunately the scope of generalization is unspecified. 
As a result, a major problem comes from the divergence between the domain 
where the model was originally trained 
and the domain where it actually operates. 

So the first requirement for the quality assurance of a DNN model is to be \emph{operational}, i.e., to 
focus on the concrete domain where the model actually operates. 
Logically speaking, the quality of a trained DNN model will be pointless without considering its operation domain.
In practice, the performance of a model may drop significantly with domain shift~\cite{li2019boosting}.
On the other hand, focusing on the operation domain also relieves the DNN model 
from depending on its original training data. 
Apart from practical concerns such as protecting the privacy and property of the training data, 
decoupling a model from its training data and process will also be helpful for (re)using it 
as a commercial off-the-shelf (COTS) software product~\cite{zhou2016learnware}. 
This viewpoint from software engineering is in contrasting to machine learning techniques dealing with domain shift such as 
transfer learning or domain adaptation that heavily rely 
on the original training data and hyperparameters~\cite{pan2009survey,shu2018a,transferlearning.xyz}. 
They need original training data because they try to \emph{generalize} 
the scope of the model to include the new operation domain.

The second requirement is to embrace the uncertainty that is intrinsic in DNN models. 
A defect, or a ``bug'', of a software artifact is a case that it does not deliver its promise. 
Different from conventional software artifacts, 
a DNN model never promises to be certainly correct on any given input, 
and thus individual incorrect predictions \emph{per se} should not be regarded as bugs, 
but to some extent features~\cite{ilyas2019adversarial}.   
Nevertheless, the model \emph{statistically} quantifies the uncertainty of their predictions. 
Collectively, it is measured with metrics such as accuracy or precision. 
Individually, it is stated by the confidence value about the prediction on each given input.  
These qualifications of uncertainty, as well as the predictions a model made, 
should be subject to quality assurance. 
For example, given a DNN model and its operation domain, 
operational testing~\cite{li2019boosting} examines 
to what degree the model's overall accuracy is degraded by the domain shift.
Furthermore, operational calibration, which is the topic of the current paper, 
identifies and fixes the misspecified confidence values on individual inputs.  

Finally, operational quality assurance should prioritize the saving of human efforts, 
which include the cost of collecting, and especially labeling, the data in the operation domain.  
The labeling of operation data often involves physical interactions, such as surgical biopsies  
and destructive testings, and thus can be expensive and time-consuming. 
Note that, as exemplified by the EU GDPR~\cite{eu-269-2014}, there are 
increasing concerns about the privacy and property rights in the data used to train DNN models. 
So when adopting a DNN model trained by a third party, 
one should not assume the availability of its original training data~\cite{zhou2016learnware, Shokri2015CCS, konevcny2016federated}. 
Without the access to the original training data, 
re-training or fine-tuning a DNN model to an operation domain can be unaffordable 
because it typically requires a large amount of labeled examples to work. 
Quality assurance activities often have to work under a much tighter budget for labeling data.

\begin{figure}[htbp]
\begin{center}
\includegraphics[width=.95\columnwidth]{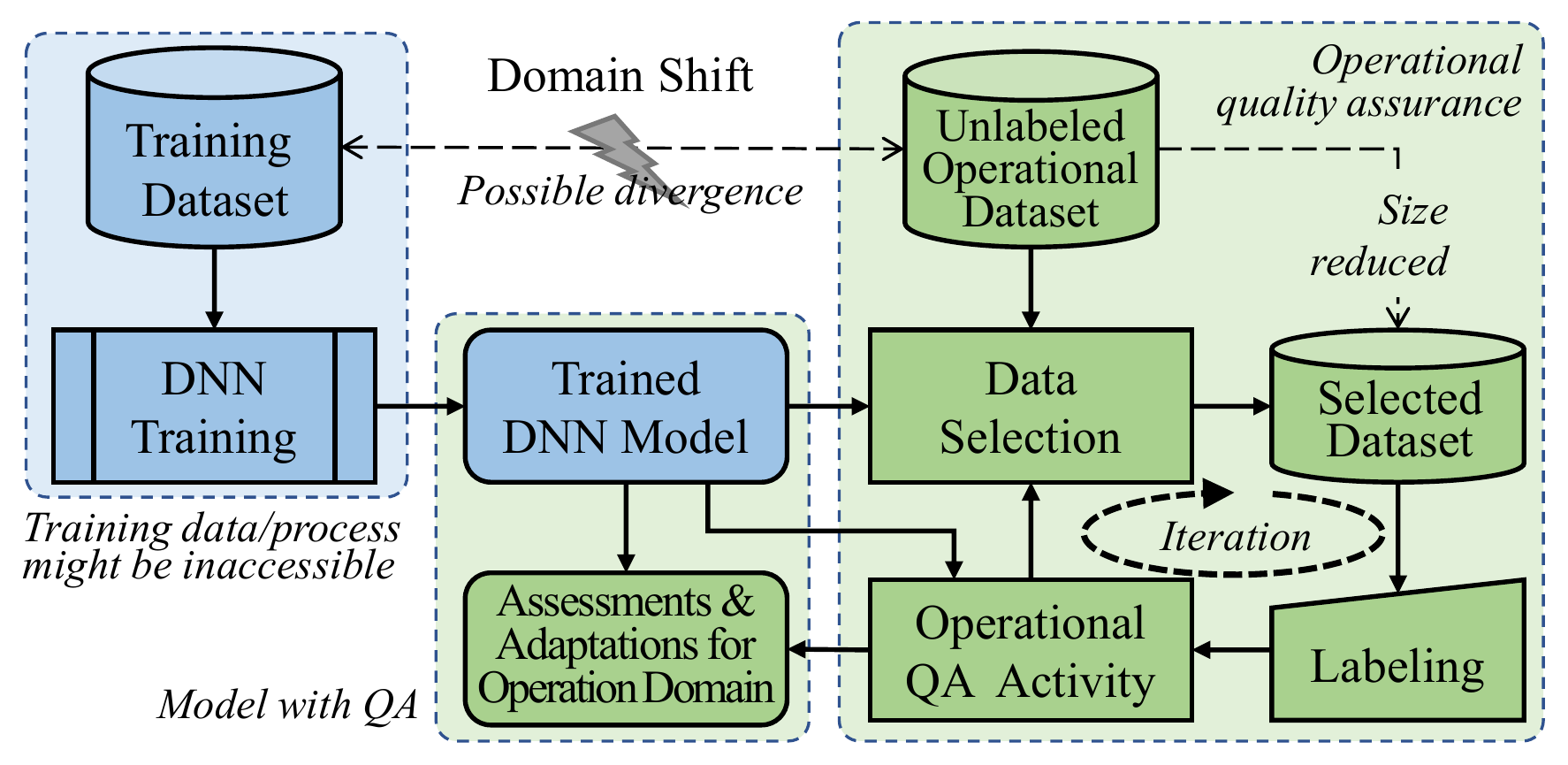}
\caption{Operational quality assurance}
\label{fig:opqa}
\end{center}
\end{figure}

Figure~\ref{fig:opqa} depicts the overall idea for operational quality assurance, 
which generalizes the process of operational testing proposed in~\cite{li2019boosting}. 
A DNN model, which is trained by a third party with the data from the origin domain, 
is to be deployed in an operation domain. It needs to be evaluated, and possibly adapted, 
with the data from the current operation domain. To reduce the effort of labeling, 
data selection can be incorporated in the procedure with the guidance of the information generated by  
the DNN model and the quality assurance activity.
Only the DNN models that pass the assessments and are possibly equipped 
with the adaptations will be put into operation.

\added{
\medskip
For example, consider a scenario that a hospital decides to equip its radiology 
department with automated medical image analysis enabled by Deep Learning~\cite{obermeyer2016predicting}. 
While the system may involve many functionalities such as clinical workflow management, 
computer-aided diagnosis, and computer-assisted reporting, the key component is a DNN model or an ensemble of DNN models 
acting as a radiologist to classify images~\cite{MCM20DLRadiology}. 
It is too expensive and technically demanding for the hospital to collect enough high quality data and 
train the model in-house. 
So the hospital purchases the model from a third-party provider, and uses it as a COTS software component. 
However, the training conditions of the model are likely to be different from the operation conditions,
and the model performance reported by the provider is unreliable due to potential issues 
such as data mismatch, selection bias, and non-stationary environments~\cite{WMD18DLbiomedicine,Zech18VarGenDLRadiograph}. 
To assure the system's quality of service, the model must be tested and calibrated against the current operation domain. 
The hospital first assesses the real accuracy of the model through operational testing~\cite{li2019boosting}, 
and decides to adopt it or not accordingly. 
Once the model is adopted and deployed, operational calibration steps in to adapt the model to the current operation domain, 
by fixing the errors in the confidence values and avoiding high-confidence false predictions. }

\added{
As a software quality assurance task, operational calibration tries to achieve best efficacy 
with a limited budget. The cost here, however, is mainly spent on the labeling of 
operation data. For sophisticated DNN models used for medical imaging classification such as Inception-V3, ResNet-50 and DenseNet-121~\cite{liu2017detecting, Zech18VarGenDLRadiograph}, 
even thousands of labeled data 	could be too less 
for model retraining or fine-tuning to work properly (c.f. Figure~\ref{fig:fine-tune1}). 
Nevertheless, as will be shown later, a deliberately designed calibration method 
can identify and fix most of the erroneous high confidence values associated to false predictions 
with only tens to small hundreds of labeled data (c.f. Figure~\ref{fig:efficiency-curve}).  
A model well calibrated for its operation domain, 
although still suffering from some loss in prediction accuracy, 
becomes more reliable in that broken promises are far less likely.   
%
}


\section{Operational calibration Problem}
\label{sec:problem}

Now we focus on the problem of operational calibration. 
We first briefly introduce DNN classifiers and their prediction confidence to pave the way for 
the formal definition of the problem. 

\subsection{DNN classifier and prediction confidence}
\label{subsec:dnnconfidence}

A deep neural network classifier contains multiple hidden layers between its input and output layers. 
A popular understanding~\cite{Goodfellow-et-al-2016} of the role of these hidden layers is that 
they progressively extract abstract features (e.g., a wheel, human skin, etc.) 
from a high-dimensional low-level input (e.g., the pixels of an image). 
These features provide a relatively low-dimensional high-level \emph{representation} $\bm{z}$ for the input $\bm{x}$, which makes the classification much easier, e.g., the image is more likely to be a car if wheels are present. 

What a DNN classifier tries to learn from the training data is a posterior probability distribution, 
denoted as $p(y \mid \bm{x})$~\cite{bishop2006prml}.
For a \textit{K}-classification problem, the distribution can be written as 
$p_i(\bm{x}) = p(y = i \mid \bm{x})$, where $i=1,2,\dots,K$. 
For each input $\bm{x}$ whose representation is $\bm{z}$, 
the output layer first computes the non-normalized prediction  
$\bm{h} =\bm{W}^{\top} \bm{z} + \bm{b}$, 
whose element $h_i$  is often called the \emph{logit} for the $i$-th class. 
The classifier then normalizes $\bm{h}$ with a softmax function to approximate the posterior probabilities
\begin{equation}
\hat{p}_i(\bm{x}) = \text{softmax}(\bm{h})_i=\frac{e^{h_i} }{\sum_{j=1}^{K} e^{h_j}}, \quad i=1, \ldots, K.
\end{equation}
%
%
Finally, to classify $\bm{x}$, one just chooses the 
the category corresponding to the maximum posterior probability, 
i.e.,  
 \begin{equation}
 \hat{y}(\bm{x}) = \mathop{\arg\max}_i \hat{p}_i(\bm{x}).
 \end{equation}

Obviously, this prediction is intrinsically uncertain.  
The \emph{confidence} for this prediction, which quantifies the  
likelihood of correctness, can be naturally measured as 
the estimated posterior class probability 
 \begin{equation}
 \hat{c}(\bm{x}) = \hat{p_i}(\bm{x}),  \quad i=\hat{y}(\bm{x}).
 \end{equation}
Confidence takes an important role in decision-making. 
For example, if the loss due to an incorrect prediction 
is four times of the gain of a correct prediction, 
one should not invest on predictions with confidence less than 0.8. 

Modern DNN classifiers are often inaccurate in confidence~\cite{szegedy2016rethinking}, 
because they overfit to the surrogate loss used in training~\cite{guo2017calibration, tewari2007consistency}. 
Simply put, they are over optimized toward the accuracy of classification, 
but not the accuracy of estimation for posterior probabilities. 
To avoid the potential loss caused by inaccurate confidence, 
\emph{confidence calibration} can be employed in the learning process~~\cite{flach2016calibration,guo2017calibration,tewari2007consistency}. 
\added{Early calibration methods such as isotonic regression~\cite{zadrozny2002transforming}, 
histogram binning~\cite{zadrozny2002transforming}, and Platt scaling~\cite{Platt1999probabilisticoutputs} simply 
train a regression model taking the uncalibrated confidence as input with the validation dataset.
More flexible methods, e.g., Temperature Scaling~\cite{hinton2015distilling}, 
find} a function $R$
to correct the logit $\bm{h}$ such that 
 \begin{equation}
 \hat{c}(\bm{x}) = \hat{p_i}(\bm{x})=\text{softmax}(R(\bm{h}))_i,  \quad i=\hat{y}(\bm{x})
 \label{equ:logitcali}
 \end{equation}
matches the real posterior probability $p_i(\bm{x})$. 
Notice that, in this setting the inaccuracy of confidence is viewed as a kind of 
systematic error or bias, not associated with particular inputs or domains.
That is, \added{the calibration does not distinguish between different inputs with the same uncalibrated confidence.} \deleted{$R$ does not take $\bm{x}$ or $\bm{z}$ as input}.

\subsection{Operational confidence calibration}

Given a domain where a previously trained DNN model is deployed, 
operational calibration 
identifies and fixes the model's errors in the confidence of predictions on individual inputs
in the domain.  
%
%
Operational calibration is conservative in that it does not change the predictions made by the model, 
but tries to give accurate estimations on the likelihood of the predictions being correct. 
With this information, a DNN model will be useful even though its prediction accuracy 
is severely affected by the domain shift. 
One may take only its predictions on inputs with high confidence, 
but switch to other models or other backup measures if unconfident.

To quantify the accuracy of the confidence of a DNN model on a dataset 
$D = \{(\bm{x}_i, y_i), i=1,\dots,N\}$, 
one can use the  
Brier score (BS)~\cite{brier1950verification}, which is actually the mean squared error of the estimation:
\begin{equation}
BS(D) = \frac{1}{N}\sum_{i=1}^N (\mathbb{I}(\bm{x}_i) - \hat{c}(\bm{x}_i))^2, 
\end{equation}
where $\mathbb{I}(\bm{x})$ 
is the indicator function for whether the labeled input $\bm{x}$ 
is misclassified or not, i.e., 
$\mathbb{I}(\bm{x}) = 1$ if $\hat{y}(\bm{x})=y(\bm{x})$, and $0$ otherwise.  

Now we formally define the problem of operation calibration:
\begin{problem} 	
Given $\mathfrak{M}$ a previously trained DNN classifier, 
$S$ a set of $N$ unlabeled examples collected from an operation domain, 
and a budget $n \ll N$ for labeling the examples in $S$, 
the task of operational calibration is to find a confidence estimation function 
$\hat{c}(\cdot)$ for $\mathfrak{M}$ with minimal Brier score $BS(S)$.
\end{problem}

Notice that operational calibration is different from the confidence calibration discussed in 
Section~\ref{subsec:dnnconfidence}.
The latter is domain-independent and usually included as a step in the training process of a DNN model (one of machine learning's focuses), 
but the former is needed when the model is deployed as a software component by a third party in a specific operation domain (what software engineering cares about). 
Technically, operational calibration cannot take the confidence error as a systematic error of the learning process, 
because the error is caused by the domain shift 
from the training data to the operation data, and it may \added{assign different confidence values to inputs with the same uncalibrated confidence value.} \deleted{depend on specific inputs from the operation domain.}


\section{Solving Operational Calibration with Gaussian Process Regression}
\label{sec:method}

At first glance operational calibration seems a simple regression problem with BS as the loss function. 
However, a direct regression would not work because of the limited budget of labeled operation data. 
It is helpful to view the problem in a Bayesian way. 
At the beginning, we have a prior belief about the correctness of a DNN model's predictions, 
which is the confidence outputs of the model.  
Once we observe some evidences that the model makes correct or incorrect predictions on some inputs, 
the belief should be adjusted accordingly.
The challenge here is to strike a balance between 
the priori that was learned from a huge training dataset but suffering from domain shift, 
and the evidence that is collected from the operation domain but limited in volume. 


\subsection{Modeling with Gaussian Process}
 \label{subsec:modeling with Gaussian Process}

It is natural to model the problem as a Gaussian Process~\cite{rasmussen2005GPML}, 
because what we need is actually a function $\hat{c}(\cdot)$. 
Gaussian Process is a non-parametric kind of Bayesian methods, 
which convert a prior over functions into a posterior over functions according to observed data.

For convenience, instead of estimating $\hat{c}(\cdot)$ directly,  
we consider  
\begin{equation}
h(\bm{x}) = \hat{c}(\bm{x}) - c_\mathfrak{M}(\bm{x}), 
\label{equ:deltaconf}
\end{equation}
where $c_\mathfrak{M}(\bm{x})$ is the original confidence output of $\mathfrak{M}$ for input $\bm{x}$.
At the beginning, without any evidence against $c_\mathfrak{M}(\bm{x})$, 
we assume that the prior distribution of $h(\cdot)$ is a zero-mean normal distribution
\begin{equation}
h \sim \mathcal{N}(\cdot \mid 0, k(\cdot, \cdot)),
\label{equ:normcov}
\end{equation}
where $k(\cdot, \cdot)$ is the covariance (kernel) function, which intuitively describes the ``smoothness'' of $h(\bm{x})$ from point to point. In other words, the covariance function ensures that $h$
produces close outputs when inputs are close in the input space.

Assume that we observe a set  of independent and identically distributed (i.i.d.) labeled operation data $\bm{I} = \{(\bm{x}_i, y_i) \mathop{|} 1\leq i \leq n\}$, in which   $y = h(\bm{x}) = \mathbb{I}(\bm{x}) - c_\mathfrak{M}(\bm{x})$. 
For notational convenience, let 
\begin{equation*}
\begin{aligned}
\bm{X} &= (\bm{x}_1^{\mathsf{T}}; \dots; \bm{x}_n^{\mathsf{T}})  \quad \text{and} \\
\bm{h} &= ({h(\bm{x}_1)}; {\dots}; {h(\bm{x}_n)})
\end{aligned}
\end{equation*}
be the observed data and their corresponding $y$-values, and let 
\begin{equation*}
\begin{aligned}
\bm{X}' &= (({\bm{x}'_{1}})^{\mathsf{T}}; \dots; ({\bm{x}'_{n}})^{\mathsf{T}}), \quad \text{and} \\
\bm{h}' &= ({h(\bm{x}'_{1})}; {\dots}; {h(\bm{x}'_{n})})
\end{aligned}
\end{equation*}
be those for a set $T=\{(\bm{x}'_i, y'_i), i=1,\dots,m\}$ of i.i.d.\ predictive points. 
We have
\begin{equation}
\left(\begin{array}{l}{\bm{h}} \\ {\bm{h'}}\end{array}\right) 
\mid \bm{X}, \bm{X}' \sim \mathcal{N}\left(\bm{0},
\left(
\begin{array}{cc}
{K_{\bm{X}\bm{X}}} & {K_{\bm{X}\bm{X}'}} \\ 
{K_{\bm{X}' \bm{X}}} & {K_{\bm{X}'\bm{X}'}}
\end{array}\right)\right)
\end{equation}
where $\bm{K}$ is the kernel matrix. 
Therefore, the conditional probability distribution is 
\begin{equation}
\bm{y'} \mid \bm{y}, \bm{X}, \bm{X}' \sim \mathcal{N}\left(\bm{\mu}', \bm{\Sigma}'\right)
\end{equation}
where
\begin{equation*}
\begin{aligned}
\bm{\mu}' &=\bm{K}_{\bm{X}' \bm{X}}\left(\bm{K}_{\bm{X} \bm{X}}\right)^{-1} \bm{y}, \\ 
\bm{\Sigma}' &=\bm{K}_{\bm{X}' \bm{X}'}-\bm{K}_{\bm{X}'\bm{X}} \left(\bm{K}_{\bm{X} \bm{X}}\right)^{-1} \bm{K}_{\bm{X} \bm{X}'}.
\end{aligned}
\end{equation*}

%

With this Gaussian Process, we can estimate the probability distribution of the operational confidence for any input $\bm{x'}$ as follows
\begin{equation}
h(\bm{x'}) \mid \bm{x'}, \bm{X}, \bm{h} \sim \mathcal{N}(\mu, \sigma),
\end{equation}
where
\begin{equation*}
\begin{aligned}
& \mu = \bm{K}_{\bm{x'} \bm{X}}(\bm{K}_{\bm{X}\bm{X}})^{-1} \bm{h}, \\
& \sigma = \bm{K}_{\bm{x'} \bm{x'}} - \bm{K}_{\bm{x'}\bm{X}}(\bm{K}_{\bm{X} \bm{X}})^{-1} \bm{K}_{\bm{X}\bm{x'}}.
\end{aligned}
\end{equation*}
Then, with Equation~\ref{equ:deltaconf}, we have the distribution of $\hat{c}\left(\bm{x'}\right)$
\begin{equation}
P\left(\hat{c}\left(\bm{x'}\right) \mid \bm{x'}\right) \sim \mathcal{N}\left(c_\mathfrak{M}\left(\bm{x'}\right) + \mu, \sigma \right).
\end{equation} 

Finally, due to the value of confidence ranges from 0 to 1, 
we need to truncate the original normal distribution~\cite{burkardt2014truncated}, i.e.,
\begin{equation}
P\left(\hat{c}\left(\bm{x'}\right) \mid \bm{x'}\right) \sim \mathcal{TN}\left(\mu_{tn}, \sigma_{tn}; \alpha, \beta \right),
\end{equation} 
where
\begin{equation}\label{truncated normal distribution}
\begin{aligned}
& \mu_{tn} = c_\mathfrak{M}\left(\bm{x'}\right) + \mu + \frac{\phi(\alpha) - \phi(\beta)}{\Phi(\alpha) - \Phi(\beta)}\sigma,  \\
& \sigma_{tn}^2 = \sigma^2 \left[1 + \frac{\alpha \phi(\alpha) - \beta \phi(\beta)}{\Phi(\beta) - \Phi(\alpha)} - \left( \frac{\phi(\alpha) - \phi(\beta)}{\Phi(\beta) - \Phi(\alpha)} \right)^2 \right], \\
& \alpha = (0 - c_\mathfrak{M}\left(\bm{x'}\right) - \mu) / \sigma,   
\quad  \beta = (1 - c_\mathfrak{M}\left(\bm{x'}\right) - \mu) / \sigma.
\end{aligned}
\end{equation}
Here the $\phi(\cdot)$ and $\Phi(\cdot)$ are the probability density function and the cumulative distribution function of standard normal distribution, respectively.

With this Bayesian approach, 
we compute a distribution, rather than an exact value, 
for the confidence of each prediction. 
To compute the Brier score, 
we simply choose the maximum a posteriori (MAP), i.e., the mode of the distribution,
as the calibrated confidence value.
Here it is the mean of the truncated normal distribution 
\begin{equation} 
	\hat{c}(\bm{x}) = \mu_{tn}.
\end{equation}



\subsection{Clustering in representation space}


Directly applying the above Gaussian Process to estimate $\hat{c}(\cdot)$ would be 
ineffective and inefficient. 
It is difficult to specify a proper covariance function in Equation~\ref{equ:normcov}, 
because the correlation between the correctness of predictions on different examples 
in the very high-dimensional input space is difficult, if possible, to model.  

Fortunately, we have the DNN model $\mathfrak{M}$ on hand, 
which can be used as a feature extractor, 
although it may suffer from the problem of domain shift
~\cite{bengio2012unsupervised}.
In this way we transform each input $\bm{x}$ from the input space to 
a corresponding point $\bm{z} $ in the representation space, 
which is defined by the output of the neurons in the last hidden layer. 
It turns out that the correctness of $\mathfrak{M}$'s predictions has 
an obvious locality, i.e., a prediction is more likely to be correct/incorrect 
if it is near to a correct/incorrect prediction in the representation space. 

Another insight for improving the efficacy and efficiency of the Gaussian Process is that 
the distribution of operation data in the sparse representation space is far from even. 
They can be nicely grouped into a small number (usually tens) of clusters,
and the correlation of prediction correctness within a group is much stronger than that between groups. 
Consequently, instead of regression with a universal Gaussian Process, 
we carry out a Gaussian Process regression in each cluster.  

This clustering does not only reduce the computational cost of the Gaussian Processes, 
but also make it possible to use different covariance functions for different clusters. 
The flexibility makes our estimation more accurate. 
Elaborately, we use the RBF kernel 
\begin{equation}
k(\bm{z}_1,\bm{z}_2) = \exp\left(-\frac{{\| \bm{z}_1 - \bm{z}_2 \|}^2}{2 \ell^2} \right)
\end{equation} 
where the 
parameter $\ell$ (length scale) can be decided according to the distribution of the original confidence produced 
by $\mathfrak{M}$.

\subsection{Considering costs in decision}
The cost of misclassification must be taken into account in real-world decision making. 
\replaced{We propose to}{One can} also measure how well a model is calibrated with the \emph{loss due to confidence error} (LCE) against a given cost model. 

For example, let us assume a simple cost model in which the gain for a correct prediction is 1 and the loss for a false prediction is $u$. The net gain if we take action on a prediction for input $\bm{x}$ will be
$ \mathbb{I}(\bm{x}) -  u \cdot (1-\mathbb{I}(\bm{x}))$. 
We further assume that there will be no cost to take no action 
when the expected net gain is negative. 
Then the actual gain for an input 
$\bm{x}$ with estimated confidence $\hat{c}(\bm{x})$ will be
\begin{equation}
g(\bm{x}) = 
\left\{
\begin{array}{cc}
{ \mathbb{I}(\bm{x}) -  u \cdot (1-\mathbb{I}(\bm{x}))} & {\text{if}~\hat{c}(\bm{x}) \ge \lambda,} \\
{0} & {\text{if}~\hat{c}(\bm{x}) < \lambda,}  
\end{array}\right.
\label{eqn:actgain}
\end{equation}
where $\lambda= \frac{u}{1+u}$ is the break-even threshold of confidence for taking action.
%
%
On the other hand, if the confidence was perfect, 
i.e., $\hat{c}(\bm{x})=1$ if the prediction was correct, and 0 otherwise, 
the total gain for dataset $D$ would be 
a constant $G_D=\sum_{i=1}^N\mathbb{I}(\bm{x}_i)$. 
So the average LCE over a dataset $D$ with $N$ examples is
$\ell(D) = \frac{1}{N}\left(G_D - \sum_{i=1}^N g(\bm{x}_i)\right)$.


With the Bayesian approach we do not have an exact $\hat{c}(\bm{x})$ 
but a truncated normal distribution of it. 
If we take $\mu_{tn}(\bm{x})$ as $\hat{c}(\bm{x})$,
the above equations still hold.%

Cost-sensitive calibration targets at minimizing the LCE instead of the Brier score. 
Notice that calibrating confidence with Brier score generally reduces LCE. 
However, with a cost model, the optimization toward minimizing LCE can be more effective and efficient. 

\subsection{Selecting operation data to label}
\label{subsec:inputselection}
In case that the set of labeled operation data is given, 
we simply apply a Gaussian Process in each cluster in the representation space 
and get the posteriori distribution for confidence $\hat{c}(\cdot)$.
However, if we can decide which operation data to label, 
we shall spend the budget for labeling more wisely. 

Initially, we select the operational input at the center of each cluster to label, 
and apply a Gaussian Process in each cluster with this central input
to compute the posterior probability distribution of the confidence. 
Then we shall select the most ``helpful'' input to label and repeat the procedure. 
The insight for input selection is twofold. 
First, to reduce the uncertainty as much as possible, 
one should choose the input with maximal variance $\sigma_{tn}^2$.
Second, to reduce the LCE as much as possible, 
one should pay more attention to those input with confidence 
near to the break-even threshold $\lambda$. 
So we chose $\bm{x}^*$ as the next input to label:
\begin{equation} \label{equ:selection}
\bm{x}^* = \arg\min_{\bm{x}} \frac{|\mu_{tn}(\bm{x}) - \lambda|}{\sigma_{tn}(\bm{x})}.
\end{equation}

\added{
With $\bm{x}^*$ and its label $y^*$, we update the corresponding Gaussian Process model and 
get better $\mu_{tn}(\cdot)$ and $\sigma_{tn}(\cdot)$. 
The select-label-update procedure is repeated until the labeling budget is used up. 
}

%

\medskip 

Putting all the ideas together, we have Algorithm~\ref{alg:1} shown below. 
The algorithm is robust in that it does not rely on any hyperparameters 
except for the number of clusters.
It is also conservative in that it does not change the predictions made by 
the model. As a result, it needs no extra validation data.

\begin{algorithm}
\renewcommand{\algorithmicrequire}{\textbf{Input:}}
\renewcommand{\algorithmicensure}{\textbf{Output:}}
\caption{Operational confidence calibration}
\label{alg:1}
\begin{algorithmic}[1]
\REQUIRE A trained DNN model $\mathfrak{M}$,  
unlabeled dataset $S$ collected from operation domain $\mathbb{D}$, 
and the budget $n$ for labeling inputs.
\ENSURE Calibrated confidence function $\hat{c}(\bm{x})$ for $\bm{x}$ belongs to $\mathbb{D}$.
\newline\emph{Build Gaussian Process models:}
\STATE Divide dataset $S$ into $L$ clusters using the K-modroid method, 
and label the inputs $\bm{o}_1,\dots,\bm{o}_L$ 
that correspond to the centers of the $L$ clusters. 
\STATE Initialize the labeled set $T = \{\bm{o}_1,\dots,\bm{o}_L\}$.
\STATE For each of the clusters, build a Gaussian Process model $gp_i$, $i=1,\dots, L$.
\WHILE {$|T| < n$}
    \STATE Select a new input $\tilde{\bm{x}} \in S \setminus T$ for labeling, where $\tilde{\bm{x}}$ is searched by Equation~\ref{equ:selection}.
    \STATE Update the Gaussian Process corresponding to the cluster containing $\tilde{\bm{x}}$.
    \STATE Update the labeled set $T \leftarrow T \cup \{\tilde{\bm{x}}\}$.
\ENDWHILE
\newline\emph{Compute confidence value for input $\bm{x}$:}
\STATE Find the Gaussian Process model $\hat{gp}$ corresponding to the cluster containing input $\bm{x}$.
\STATE Compute $\mu_{tn}(\bm{x})$ according to Equation~\ref{truncated normal distribution}.
\STATE Output the estimated calibrated confidence $\hat{c}(\bm{x}) = \mu_{tn}(\bm{x})$.
\end{algorithmic}
\end{algorithm}

\subsection{Discussions}
\label{subsec:discussions}
To understand why our approach is more effective than conventional confidence calibration techniques,
one can consider the three-part decomposition of the Brier score~\cite{murphy1973BSPartition}
\begin{equation}
\begin{aligned}
BS = &\sum_{m=1}^{M} \frac{|D_{m}|}{N} \left(\text{conf}(D_m) -\text{acc}(D_m) \right)^{2} \\
&-\sum_{m=1}^{M} \frac{1}{N} \left(\text{acc}(D_m) - \text{acc}\right)^{2}+\text{acc}(1-\text{acc}),
\end{aligned}
\end{equation}
where $D_m$ is the set of inputs whose confidence falls into 
the interval $I_{m}=\left(\frac{m-1}{M}, \frac{m}{M}\right]$, 
and the $\text{acc}(D_m)$ and $\text{conf}(D_m)$ 
are the expected accuracy and confidence in $D_m$, respectively.
The $\text{acc}$ is the accuracy of dataset $D$.

In this decomposition, the first term is called \emph{reliability}, 
which measures the distance between the confidence and the true posterior probabilities. 
The second term is \emph{resolution}, which measures the distinctions of the predictive probabilities.
The final term is \emph{uncertainty}, which is only determined by the accuracy.

In conventional confidence calibration,
the model is assumed to be well trained and work well with the accuracies.
In addition, the grouping of $D_m$ is acceptable because the confidence error is regarded as \emph{systematic error.} 
So one only cares about minimizing the reliability. 
This is exactly what conventional calibration techniques such as Temperature Scaling are designed for. 

However, in operational calibration, the model itself suffers from the domain shift, 
and thus may be less accurate than expected. 
Even worse, the grouping of $D_m$ is problematic because the confidence error is \emph{unsystematic} and 
the inputs in $D_m$ are not homogeneous anymore. 
Consequently, we need to maximize the resolution and minimize the reliability at the same time. 
Our approach achieves these two goals with  
 more discriminative calibration that is based on the features of individual inputs rather than their logits or confidence values.


This observation also indicates that the benefit of our approach over Temperature Scaling 
will diminish if the confidence error happens to be systematic. 
For example, in case that the only divergence of the data in the operation domain 
is that some part of an image is missing, our approach will perform similarly to 
or even slightly worse than Temperature Scaling. 
However, as can be seen from later experiments, most operational situations have more or less 
domain shifts that Temperature Ccaling cannot handle well.

In addition, when the loss for false prediction $u$ is very small ($u \leq 0.11$, as observed from experiments in the next section), our approach will be ineffective in reducing LCE.
It is expected because in this situation one should accept almost all predictions, 
even when their confidence values are low.

\section{Empirical evaluation}
\label{sec:evaluation}

We conducted a series of experiments to answer the following questions:
\begin{enumerate}
\item Is our approach to operational calibration generally effective in different tasks?
\item How effective it is, compared with alternative approaches?
\item How efficient it is, in the sense of saving labeling efforts? 
\end{enumerate}

We implemented our approach on top of the PyTorch 1.1.0 DL framework.
The code, together with the experimental data, are available at \url{https://github.com/Lizn-zn/Op-QA}. 
The experiments were conducted on a GPU server 
with two Intel Xeon Gold 5118 CPU @ 2.30GHz,
400GB RAM, and 10 GeForce RTX 2080 Ti GPUs.
The server ran Ubuntu 16.04 with GNU/Linux kernel 4.4.0. 

The execution time of our operational calibration 
depends on the size of the dataset used, and the architecture of the DNN model.  
For the tasks listed below, the execution time varied from about 3.5s to 50s, 
which we regard as totally acceptable. 

\subsection{Experimental tasks}
\begin{figure}[htbp]
\begin{center}
\subfloat[CIFAR-10 (origin domain)]{
\includegraphics[width=.42\columnwidth]{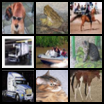}}
\hspace{2em}
\subfloat[STL-10 (operation domain)]{
\includegraphics[width=.42\columnwidth]{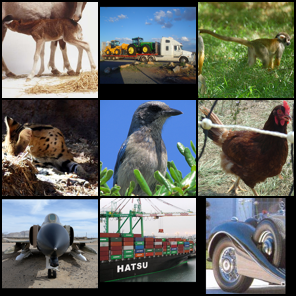}}

\subfloat[ImageCLEF-(c) (origin domain)]{
\includegraphics[width=.21\columnwidth]{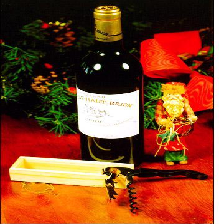}
\includegraphics[width=.21\columnwidth]{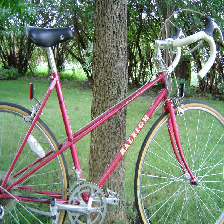}}
\hspace{2em}
\subfloat[ImageCLEF-(p) {\tiny (operation domain)}]{
\includegraphics[width=.21\columnwidth]{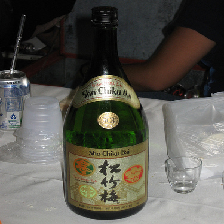}
\includegraphics[width=.21\columnwidth]{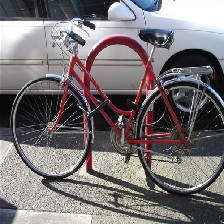}} \\
\caption{Examples of origin and operation domains. 
\added{\rm{In task~3 a ResNet-18 model was trained with low-resolution images (a), but applied to high-resolution images (b). In task~5 a ResNet-50 model was applied to images (d) with backgrounds and styles different from training images (c). }}
}
\label{fig:orig-and-opt}
\end{center}
\end{figure}

To evaluate the general efficacy of our approach, 
we designed six tasks that were different in the application domains 
(image recognition and natural language processing), 
operation dataset size (from hundreds to thousands),
classification difficulty (from 2 to 1000 classes),
and model complexity (from $\sim$$10^3$ to $\sim$$10^7$ parameters).
To make our simulation of domain shifts realistic, 
in four tasks we adopted third-party operation datasets often used in the transfer learning research,
and the other two tasks we used mutations that are also frequented made in the machine learning community.  
Figure~\ref{fig:orig-and-opt} demonstrates some example images from the origin and operation domains 
\added{for tasks 3 and 5}.
%
Table~\ref{tab:ExpDesign} lists the settings of the six tasks. 


\begin{table}[htb]
\caption{Dataset and model settings of tasks}
\centering
\setlength{\tabcolsep}{.25em}
\begin{threeparttable}
\begin{tabular}{|c|c|c|c|c|p{0.4cm}<{\centering}|}
\hline
\multirow{2}*{\textbf{Task}} & \multirow{2}*{\textbf{Model}} & \multicolumn{3}{c|}{\textbf{Origin Domain} $\rightarrow$ \textbf{Operation Domain}} \\
\cline{3-5}
& & \textbf{Dataset} & \textbf{Acc. (\%)} & \textbf{Size$^*$} \\
\hhline{|=====|}
\multirow{2}*{1} & \multirow{2}*{LeNet-5} & Digit recognition & \multirow{2}*{96.9 $\rightarrow$ 68.0} & \multirow{2}*{900} \\
& & (MNIST $\rightarrow$ USPS) &  & \\
\hline 
\multirow{2}*{2} & \multirow{2}*{RNN} & Polarity & \multirow{2}*{99.0 $\rightarrow$ 83.4} & \multirow{2}*{1,000} \\
& & (v1.0 $\rightarrow$ v2.0) & & \\
\hline
\multirow{2}*{3} & \multirow{2}*{ResNet-18} & Image classification & \multirow{2}*{93.2 $\rightarrow$ 47.1} & \multirow{2}*{5,000} \\
& & CIFAR-10 $\rightarrow$ STL-10 &  & \\
\hline
\multirow{2}*{4} & \multirow{2}*{VGG-19} & CIFAR-100 & \multirow{2}*{72.0 $\rightarrow$ 63.6} & \multirow{2}*{5,000} \\
& & (orig. $\rightarrow$ crop) &  & \\
\hline
\multirow{2}*{5} & \multirow{2}*{ResNet-50} & ImageCLEF & \multirow{2}*{99.2 $\rightarrow$ 73.2} & \multirow{2}*{480} \\
& & (c $\rightarrow$ p) &  & \\
\hline
\multirow{2}*{6} & \multirow{2}*{Inception-v3} & ImageNet & \multirow{2}*{77.5 $\rightarrow$ 45.3} & \multirow{2}*{5,000} \\
& & (orig. $\rightarrow$ down-sample) &  & \\
\hline
\end{tabular}
 \begin{tablenotes}
        \footnotesize
        \item[*] It refers to the maximum number of operation data available for labeling.
      \end{tablenotes}
\end{threeparttable}
\label{tab:ExpDesign}
\end{table}

In Task 1 we applied a LeNet-5 model originally trained 
with the images from the \textbf{MNIST} dataset~\cite{lecun1998gradient} 
to classify images from the \textbf{USPS} dataset~\cite{friedman2001elements}. 
Both of them are popular handwritten digit recognition datasets consisting of 
single-channel images  of size 16$\times$16$\times$1, \added{but the latter is more difficult to read}.
%
The size of the training dataset was 2,000, and the size of the operation dataset was 1,800.
We reserved 900 of the 1,800 operation data for testing, and used the other 900 for operational calibration. 

Task 2 was focused on natural language processing. 
\textbf{Polarity} is a dataset for sentiment-analysis~\cite{Pang2002sentiment}. 
It consists of sentences labeled with corresponding sentiment polarity (i.e., positive or negative). 
We chose Polarity-v1.0, which contained 1,400 movie reviews collected in 2002, as the training set.
The Polarity-v2.0, which contained 2,000 movie reviews collected in 2004, was used as the data from 
the operation domain. We also reserved half of the operation data for testing. 

In Task 3 we used two classic image classification datasets \textbf{CIFAR-10}~\cite{krizhevsky2009learning}
 and \textbf{STL-10}~\cite{coates2011analysis}.
The former consists of 60,000 32$\times$32$\times$3 images in 10 classes, 
and each class contains 6,000 images.
The latter has only 1,3000 images, 
but the size of each image is 96$\times$96$\times$3.
We used the whole CIFAR-10 dataset to train the model. 
The operation domain was represented by 8,000 images collected from STL-10, 
in which 5,000 were used for calibration, and the other 3,000 were reserved for testing.

Tasks 4 used the dataset \textbf{CIFAR-100}, which was more difficult than 
CIFAR-10 and contained 100 classes with 600 images in each.
We trained the model with the whole training dataset of 50,000 images.
To construct the operation domain, we randomly cropped the remaining 10,000 images. 
One half of these cropped images were used for calibration and the other half for testing.

Task 5 used the image classification dataset from the \textbf{ImageCLEF} 2014 challenge~\cite{Mller2010ImageCLEF}. 
It is organized with 12 common classes derived from three different domains: 
ImageNet ILSVRC 2012 (i), Caltech-256 (c), and Pascal VOC 2012 (p).
We chose the dataset (c) as the origin domain and dataset (p) as the operation domain.
Due to the extremely small size of the dataset, we divided the dataset (p) 
for calibration and testing by the ratio 4:1.
 
Finally, Task 6 dealt with an extremely difficult situation. 
\textbf{ImageNet} is a large-scale image classification dataset 
containing more than 1.2 million 224$\times$224$\times$3 images across 1,000 categories~\cite{Deng09imagenet}.
The pre-trained model Inception-v3 was adopted for evaluation.
The operation domain was constructed by down-sampling 10,000 images from the original test dataset.
Again, half of the images were reserved for testing.

\subsection{Efficacy of operational calibration}

Table~\ref{tab:brier-score} gives the Brier scores of the confidence before (col. Orig.) and after (col. GPR) operational calibration. 
In these experiments all operation data listed in Table~\ref{tab:ExpDesign} (not including the reserved test data) were labeled and used in the calibration.
The result unambiguously confirmed the general efficacy of our approach. 
In the following we elaborate on its relationship with the fine-tuning technique often employed in practice.  

\begin{table*}[htb]
\caption{Brier scores of different calibration methods}
\centering
\begin{threeparttable}
\begin{tabular}{|p{0.6cm}<{\centering}|c|c|c|c|c|c|c|c|c|c|c|c|}
\hline
\multirow{2}*{\textbf{Task}} & \multirow{2}*{\textbf{Model}} & \multirow{2}*{\textbf{Orig.}} & \multicolumn{3}{c|}{\textbf{Operational calibration}} & \multicolumn{4}{c|}{\textbf{Conventional calibration}} & \multirow{2}*{\textbf{SAR}}  \\
\cline{4-10} 
  & & & \textbf{GPR} & \textbf{RFR} & \textbf{SVR} & \textbf{TS} & \added{\textbf{PS-conf.}} & \added{\textbf{PS-logit}} & \added{\textbf{IR}} &   \\
\hhline{|===========|}
1  & LeNet-5 & 0.207 & \textbf{0.114} & 0.126 & 0.163 & 0.183 & \added{0.316}  & \added{0.182} & \added{0.320} & 0.320 \\
\hline
2  & RNN & 0.203 & \textbf{0.102} & 0.107 & 0.202 & 0.185 & \added{0.641} & \added{0.125} & \added{0.655} & 0.175 \\
\hline 
3  & ResNet-18 & 0.474 & \textbf{0.101} & 0.121 & 0.115 & 0.387 & \added{0.471} & \added{0.254} & \added{0.529} & 0.308 \\
\hline
4  & VGG-19 & 0.216 & \textbf{0.158} & 0.162 & 0.170 & 0.217 & \added{0.359}  & \added{0.253} & \added{0.364}  & 0.529 \\
\hline
5  & ResNet-50 & 0.226 & \textbf{0.179} & 0.204 & 0.245 & 0.556 & \added{0.789}  & \added{0.319} & \added{0.925}  & 0.364 \\
\hline
6  & Inception-v3 & 0.192 & \textbf{0.161} & 0.167 & 0.217 & 0.191 & \added{0.546} & \added{0.427} &  \added{0.334} & - \\
\hline
\end{tabular}
 \begin{tablenotes}
        \footnotesize
        \item 	Orig.--Before calibration. 
        		GPR--Gaussian Process-based approach.
				RFR--Random Forest Regression in the representation space.
				SVR--Support Vector Regression in the representation space.
				TS--Temperature Scaling~\cite{guo2017calibration}.
				\added{PS--Platt Scaling~\cite{Platt1999probabilisticoutputs}.
				IR--Isotonic Regression Scaling~\cite{zadrozny2002transforming}.
				Conf./Logit indicates that the calibration took confidence value/logit as input. }
				SAR--Regression with Surprise values~\cite{kim2019guiding}.
				We failed to evaluate SAR on task 6 because it took too long to run on the huge dataset.
  \end{tablenotes}
\end{threeparttable}
\label{tab:brier-score}
\end{table*}

\subsubsection{Calibration when fine-tuning is ineffective}

\begin{figure*}[htbp]
\begin{center}
\subfloat[Task 4]{
\includegraphics[width=0.62\columnwidth]{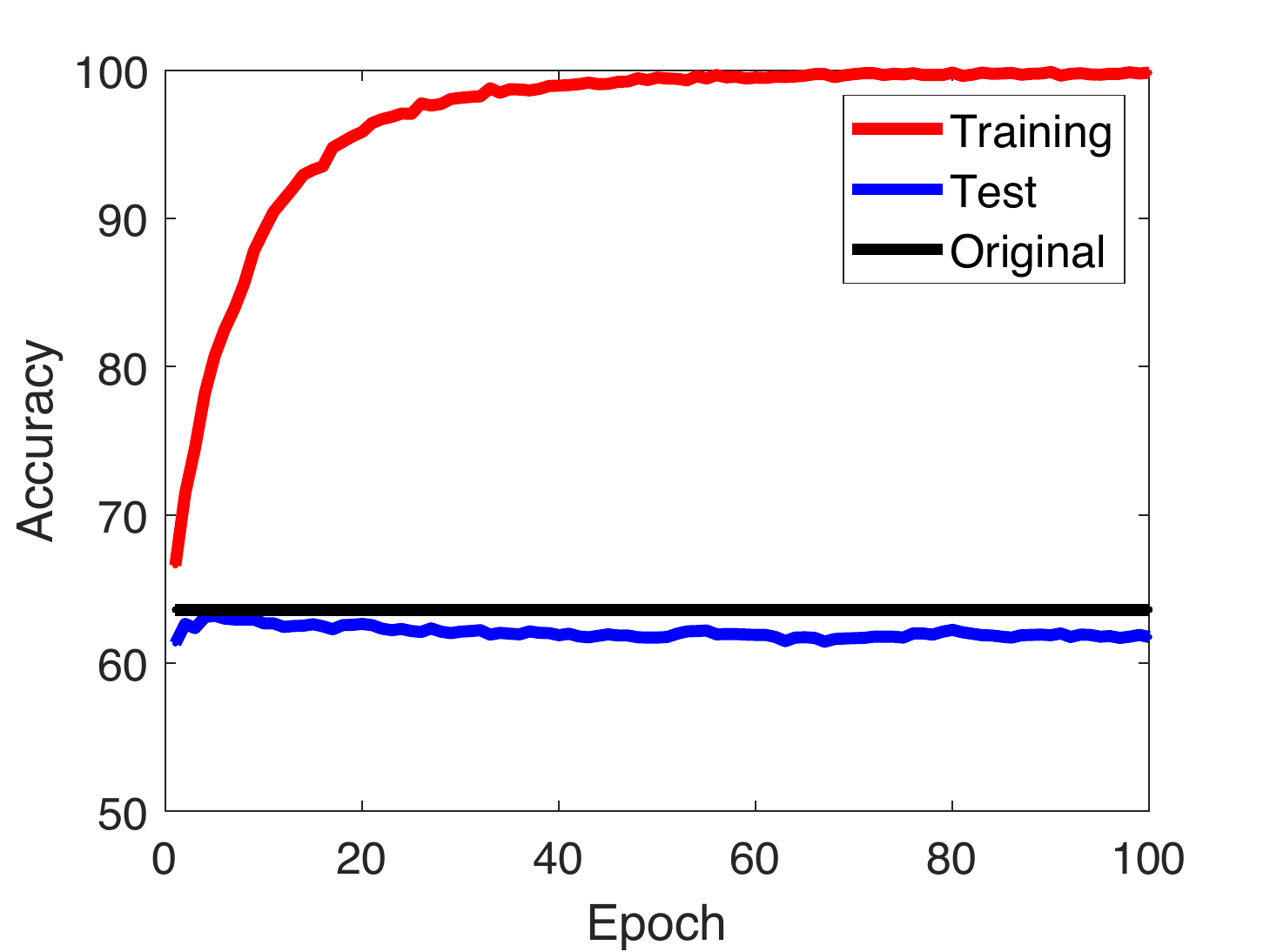}}
\subfloat[Task 5]{
\includegraphics[width=0.62\columnwidth]{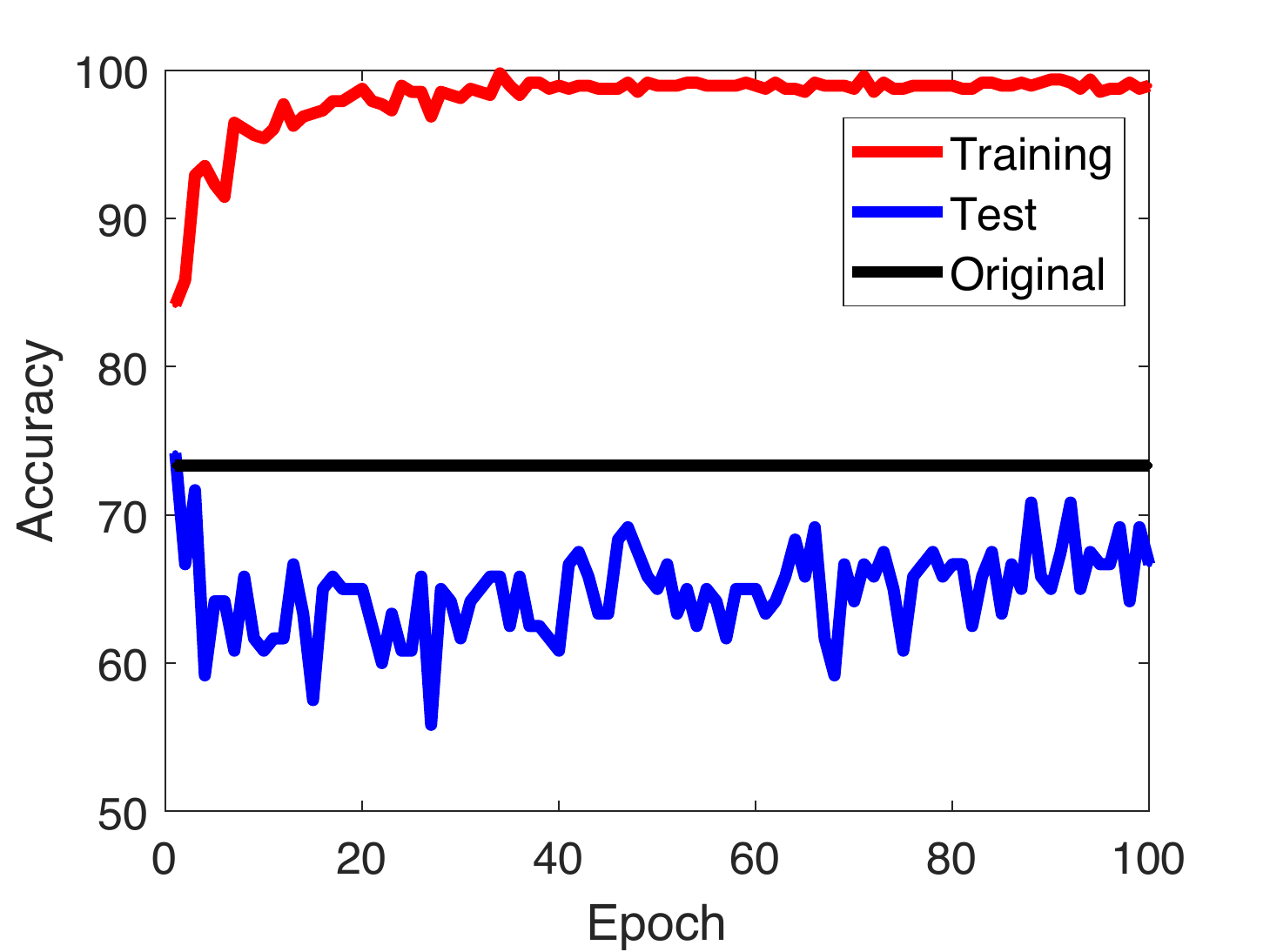}}
\subfloat[Task 6]{
\includegraphics[width=0.62\columnwidth]{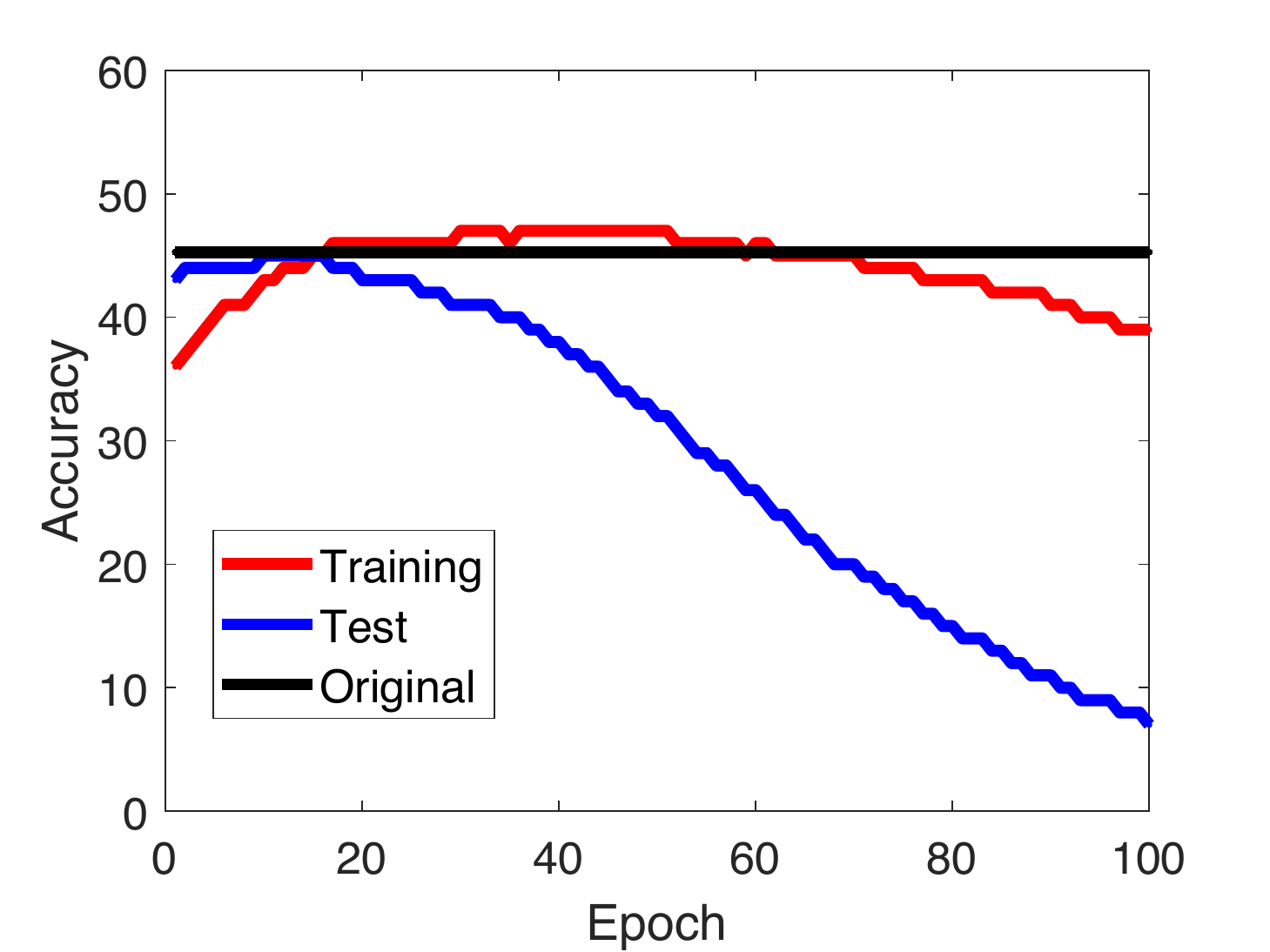}}
\caption{Ineffective fine-tuning of difficult tasks.
\added{\rm{All available operation data (5000, 480, and 5000, respectively) were used. }}}
\label{fig:fine-tune1}
\end{center}
\end{figure*}

A machine learning engineer might first consider to apply 
fine-tuning tricks to deal with the problem of domain shift. 
However, for non-trivial tasks, such as our tasks 4, 5, and 6,
it can be very difficult, if possible, to fine-tune the DNN model 
with small operation datasets. 
Figure~\ref{fig:fine-tune1} shows the vain effort in fine-tuning 
the models with all the operation data (excluding test data). 
We tried all tricks including data augmentation, weight decay, and regularization 
to avoid over-fitting but failed to improve the test accuracy.  

\begin{figure}[htbp]
\begin{center}
\includegraphics[width=0.70\columnwidth]{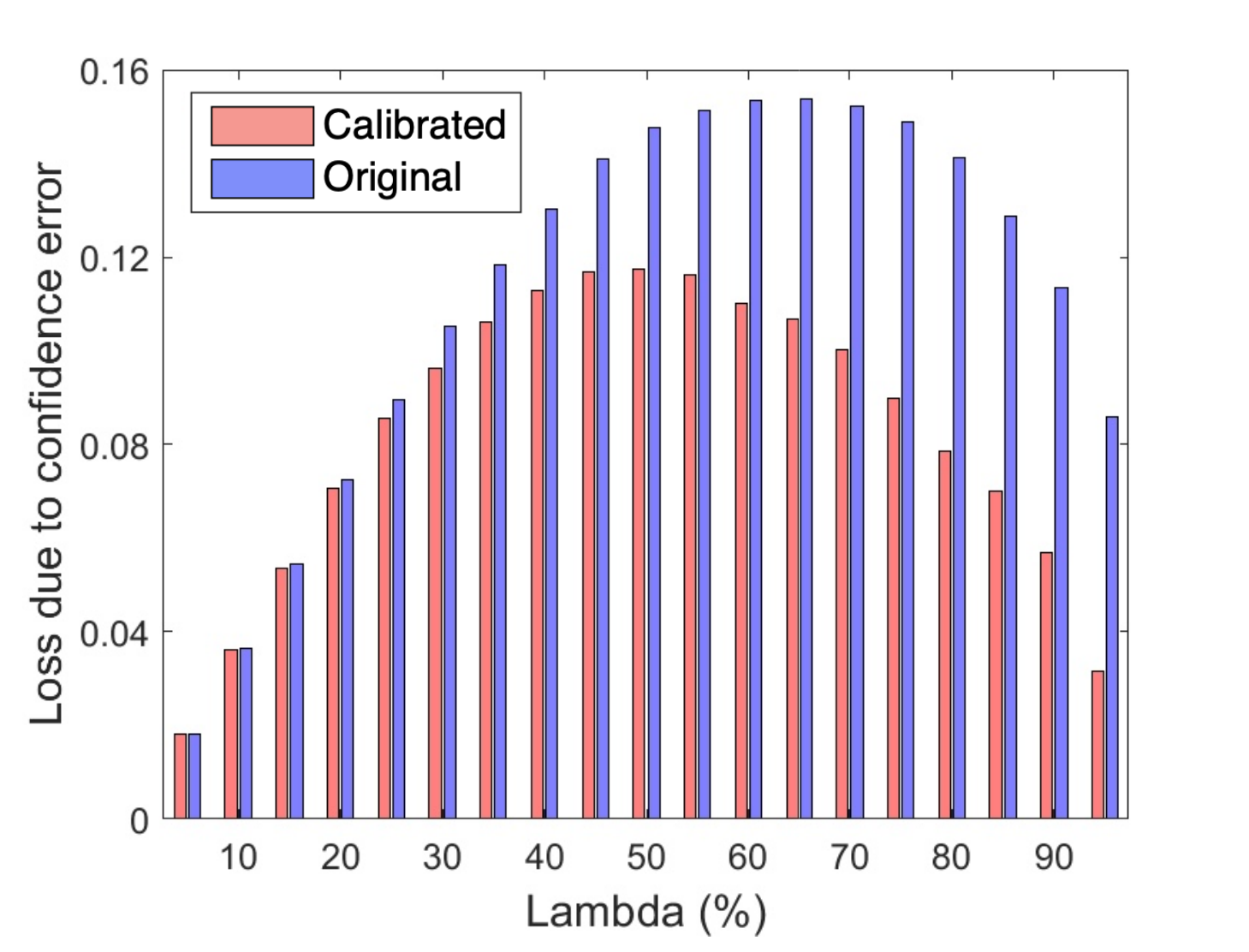}
\caption{loss due to confidence error}
\label{fig:LCE-1}
\end{center}
\end{figure}

Fortunately, our operational calibration worked quite well in these difficult situations. 
In addition to the improvement in Brier scores reported in Table~\ref{tab:brier-score}, 
we can also see the saving of LCE for task~4 in Figure~\ref{fig:LCE-1} as an example. 
Our approach reduced about a half of the LCE when $\lambda > 0.8$, 
which indicates its capability in reducing high confidence errors.

\subsubsection{Calibration when fine-tuning is effective}

In case of easier situations that fine-tuning works, 
we can still calibrate the model to give more accurate confidence. 
Note that effective fine-tuning does not necessarily provide accurate confidence. 
One can first apply fine-tuning until test accuracy does not increase, 
and then calibrate the fine-tuned model with the rest operation data.

\begin{figure}[htbp]
\begin{center}
\subfloat[\added{Task 2}]{
\includegraphics[width=0.5\columnwidth]{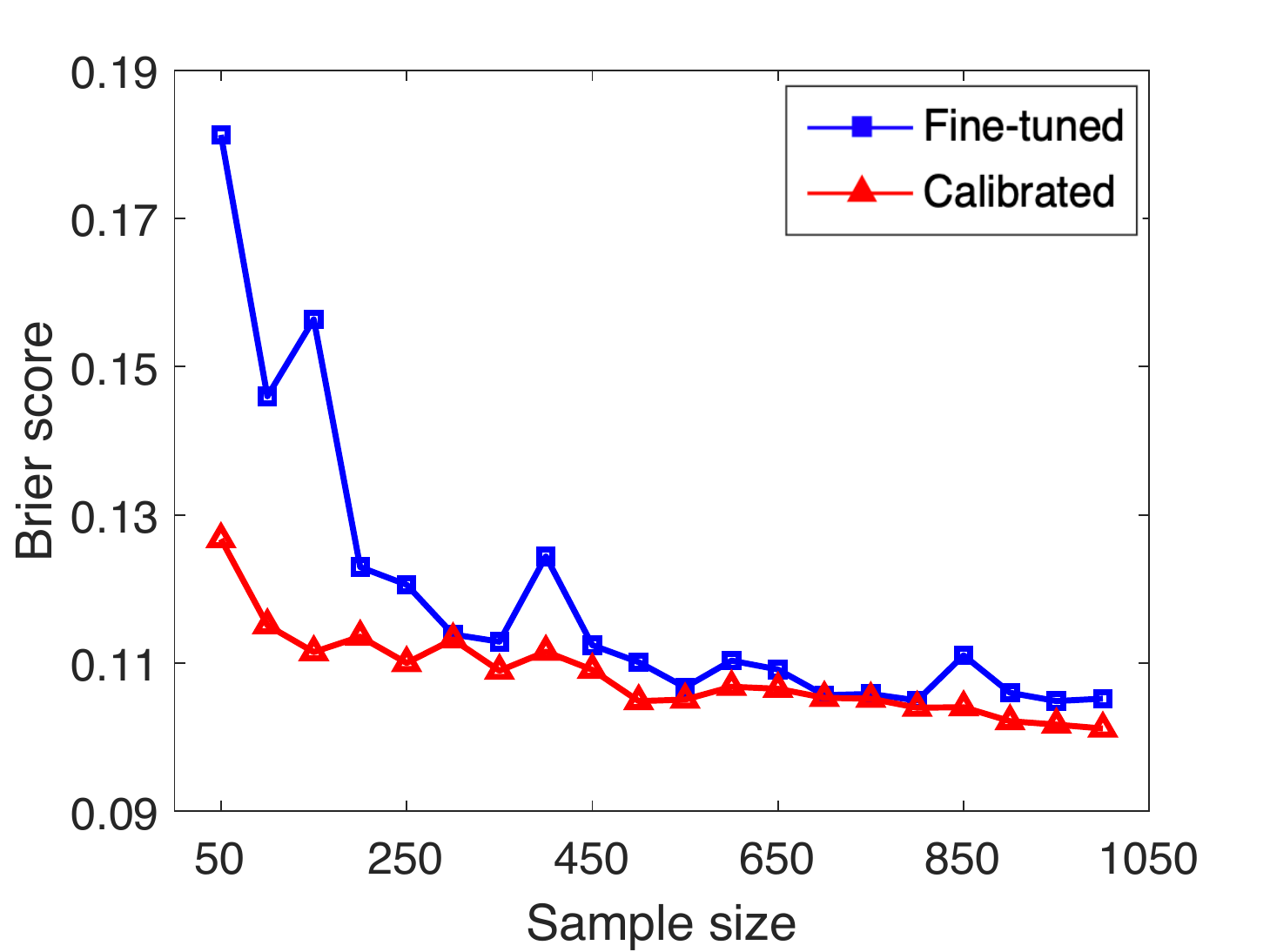}
\label{fig:ft-cali2}}
\subfloat[\added{Task 3}]{
\includegraphics[width=0.5\columnwidth]{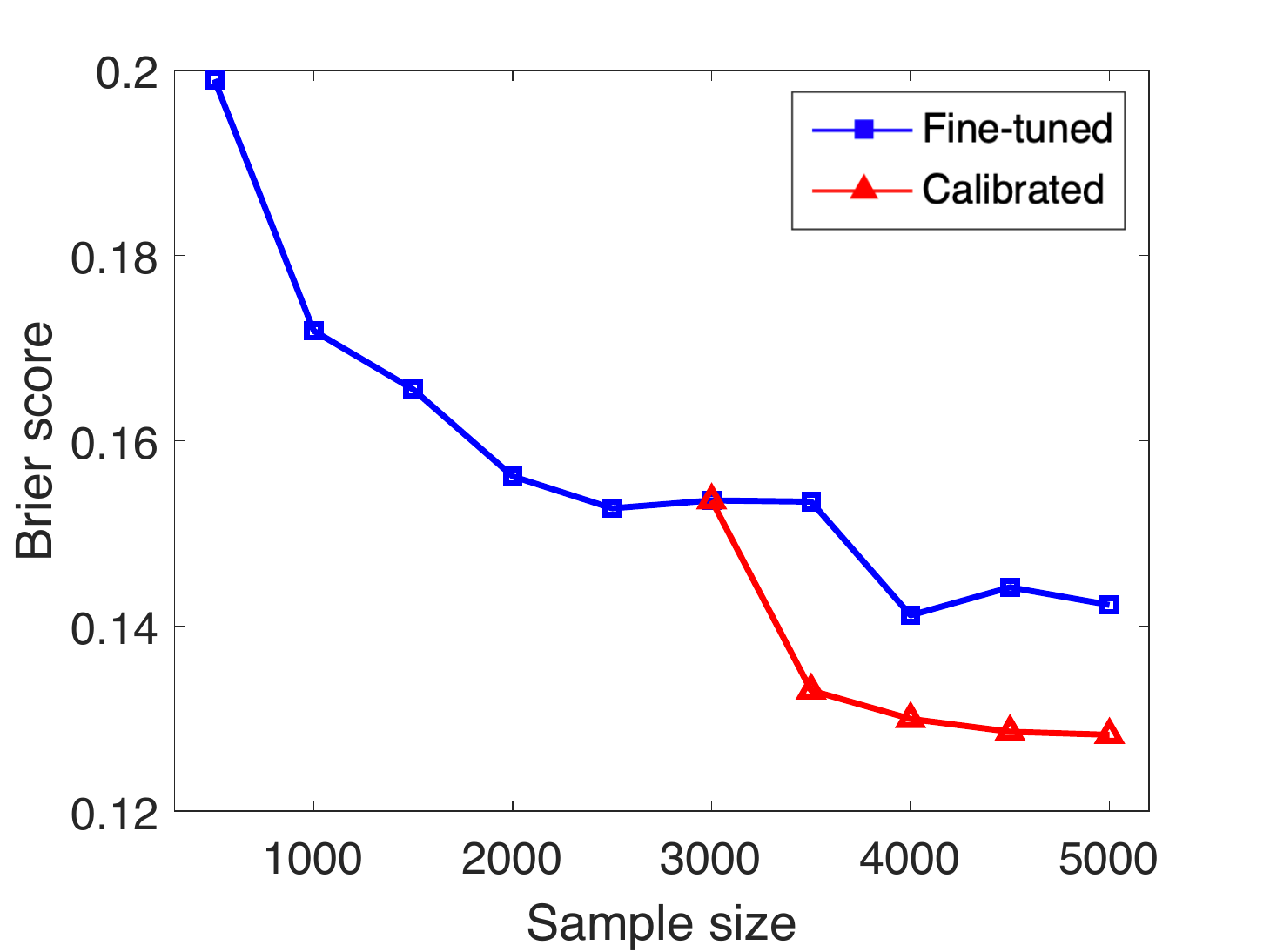}
\label{fig:ft-cali3}}
\caption{Calibration when fine-tuning is effective}
\end{center}
\end{figure}

For example, we managed to fine-tune the models in our tasks 1, 2, and 3.~\footnote{%
Here we used some information of the training process, such as the learning rates, weight decays and training epochs. Fine-tuning could be more difficult because these information could be unavailable in real-world operation settings.  
} 
Task 1 was the easiest to fine-tune and its accuracy kept increasing 
and exhausted all the 900 operational examples.
Task 2 was binary classification, in this case our calibration was actual 
an effective fine-tuning technique. 
Figure~\ref{fig:ft-cali2} shows that our approach was 
more effective and efficient than conventional fine-tuning as it converged more quickly. 
For task 3 with fine-tuning the accuracy stopped increasing at about 79\%, 
with about 3,000 operational examples.  
Figure~\ref{fig:ft-cali3} show that, 
the Brier score would decrease more if we spent rest operation data on calibration 
than continuing on the fine-tuning.

%

Based on the significant (16.1\%-78.6\%) reductions in Brier scores in all of the tasks reported in Table~\ref{tab:brier-score} and the above discussions,
we conclude that \emph{the Gaussian Process-based approach to operational calibration
is generally effective, 
and it is worthwhile no matter whether the fine-tuning works or not}.

\subsection{Comparing with alternative methods}

\added{First, we applied three widely used calibration methods, \emph{viz.} Temperature Scaling (TS)~\cite{hinton2015distilling}, 
Platt Scaling (PS)~\cite{Platt1999probabilisticoutputs}, and Isotonic Regression (IR)~\cite{zadrozny2002transforming},
with the same operation data used in our approach.
TS defines the calibration function $R$ in Equation~\ref{equ:logitcali} as $R(\bm{h}) = \bm{h}/T$,  
where $T$ is a scalar parameter computed by minimizing the 
negative log likelihood~\cite{hastie2009elements}
on a validation dataset. 
PS and IR directly work on the confidence values. 
PS trains a one-dimensional logistic regression and calibrates confidence 
as $\hat{c} = 1/(1+e^{-a\cdot c_{\mathfrak{M}}+b})$, 
where $a, b$ are scalar parameters computed by minimizing the cross entropy on a validation dataset.
IR simply fits a monotonic confidence calibration function minimizing 
the Brier score on a validation dataset. 
Note that in our experiment, the operation data were used instead of the validated dataset. 
We implemented TS according to~\citet{guo2017calibration}. For PS and IR we used the
well-known machine learning library scikit-learn~\cite{scikit-learn}.
}

As shown in Table~\ref{tab:brier-score}, TS, 
although reported to be usually the most effective conventional confidence calibration method~\cite{guo2017calibration},
was hardly effective in these cases. 
It even worsened the confidence in tasks 4 and 5. 
We observed that its bad performance  
came from the significantly lowered resolution part of the Brier score, 
which confirmed the analysis in Section~\ref{subsec:discussions}.
For example, in task 3, with Temperature Scaling the reliability decreased from 0.196 to 0.138, but the resolution dropped from 0.014 to 0.0. 
In fact, in this case the calibrated confidence values were all very closed to 0.5 after scaling. 
However, with our approach the reliability decreased to 0.107,
and the resolution also increased to 0.154. 
\added{The same reason also failed PS (col. PS-Conf.) and IR (col. IR).}

\added{We also included in our comparison an improved version of PS, which built 
a regression over the logit instead of the confidence~\cite{guo2017calibration}. 
Its calibration function $R$ for Equation~\ref{equ:logitcali} was
$R(\bm{h}) = \bm{W}^\mathsf{T} \bm{h} + \bm{b}$,
where $\bm{W}$  and $\bm{b}$ were 
computed by minimizing the cross entropy on the operation data. 
It performed much better than the original PS but 
still failed in tasks 4, 5 and 6.
}


Second, we also tried to calibrate confidence based on the Surprise value 
that measured the difference in DL system's behavior 
between the input and the training data~\cite{kim2019guiding}.
We thought it could be effective because it also leveraged the distribution of examples 
in the representation space.
We made polynomial regression between the confidence adjustments and the likelihood-based Surprise values. 
Unfortunately, it did not work for most of the cases (col. SAR in Table~\ref{tab:brier-score} ). 
We believe the reason is that Surprise values are scalars 
and cannot provide enough information for operational calibration.  

Finally, to examine whether Gaussian Process Regression (GPR) is the right choice for our operational calibration framework, 
we also experimented with two standard regression methods, viz. 
Random Forest Regression (RFR) and Support Vector Regression (SVR). 
We used linear kernel for SVR and ten decision trees for RFR.
As shown in Table~\ref{tab:brier-score}, in most cases, the non-liner RFR performed better than the linear SVR, 
and both of them performed better than Temperature Scaling but worse than GPR.
The result indicates that (1) calibration based on the features extracted by the model 
rather than the logits computed by the model is crucial, 
(2) the confidence error is non-linear and unsystematic,   
and (3) the Gaussian Process as a Bayesian method can provide better estimation of the confidence.

In summary, \replaced{our GPR approach achieved significant Brier Score reduction and outperformed 
conventional calibration methods and the Surprise value-base regression in all the tasks.}{
the Brier score reductions of our GPR approach were       
at least 3.87 times of that of Temperature Scaling and 2.24 times of Surprise value-base regression.} 
It also outperformed alternative implementations based on RFR and SVR. 
So we conclude that 
 \emph{our approach is more effective than Temperature Scaling and 
other alternative choices for operational calibration}.

\subsection{Efficiency of operational calibration}

In the above we have already shown that our approach worked with 
small operation datasets that were insufficient for fine-tuning (Task 4, 5, and 6). 
In fact, the Gaussian Process-based approach has a nice property 
that it starts to work with very few labeled examples. 
We experimented the approach with the input selection method presented in Section~\ref{subsec:inputselection}.
We focused on the number of high-confidence false predictions, which 
was decreasing as more and more operational examples were labeled and used.  

We experimented with all the tasks but labeled only 10\% of the operation data. 
Table~\ref{tab:efficiency-results} shows the numbers of high-confidence false predictions 
before and after operational calibration. 
As a reference, we also include the numbers of high-confidence correct predictions. 
We can see that most of the high-confidence false predictions were eliminated. 
It is expected that there were less high-confidence correct predictions after calibration, 
because the actual accuracy of the models dropped. 
The much lowered LCE scores, which took into account 
both the loss in lowering the confidence of correct predictions and the gain in lowering the confidence of false predictions, 
 indicate that the overall improvements were significant.  

\begin{table}[htb]
\caption{Reducing high-confidence false predictions with 10\% operation data labeled}
\centering
\setlength{\tabcolsep}{.38em}
\begin{threeparttable}
\begin{tabular}{|p{0.4cm}<{\centering}|p{1.2cm}<{\centering}|c|c|c|c|c|}
\hline
No. & Model & $\lambda$ & Correct pred. & False pred. & LCE \\
\hhline{|======|}
\multirow{2}*{1} & \multirow{2}*{LeNet-5} & 0.8 & 473 $\rightarrow$ 309.1 & \textbf{126$\rightarrow$24.3} & 0.143$\rightarrow$0.089  \\
&  &  0.9  & 417 $\rightarrow$ 141.9 & \textbf{74 $\rightarrow$ 2.5} & 0.096$\rightarrow$0.055 \\
\hline
\multirow{2}*{2} & \multirow{2}*{RNN} & 0.8 & 512 $\rightarrow$ 552.9 & \textbf{118$\rightarrow$39.9} & 0.162$\rightarrow$0.091  \\
&  &  0.9  & 482 $\rightarrow$ 261.3 & \textbf{106 $\rightarrow$12.0} & 0.132$\rightarrow$0.070 \\
\hline
\multirow{2}*{3} & {ResNet} & 0.8 & 1350 $\rightarrow$ 839.2 & \textbf{1372$\rightarrow$59.7} & 0.370$\rightarrow$0.054  \\
& {-18}  &  0.9  & 1314 $\rightarrow$ 424.0 & \textbf{1263 $\rightarrow$9.4} & 0.358$\rightarrow$0.041 \\
\hline
\multirow{2}*{4} & \multirow{2}*{VGG-19} & 0.8 & 1105 $\rightarrow$ 392.5 & \textbf{583$\rightarrow$46.9} & 0.127$\rightarrow$0.070 \\
& &  0.9  & 772$\rightarrow$142.8  & \textbf{280$\rightarrow$9.3} & 0.074$\rightarrow$0.038 \\
\hline
\multirow{2}*{5} & {ResNet} & 0.8 & 53 $\rightarrow$ 26.9 & \textbf{16$\rightarrow$5.2} & 0.162$\rightarrow$0.136  \\
& {-50} &  0.9  & 46 $\rightarrow$ 26.9 &\textbf{10$\rightarrow$2.0} & 0.108$\rightarrow$0.064 \\
\hline
\multirow{2}*{6} & {Inception} & 0.8 & 1160$\rightarrow$692.0 & \textbf{265$\rightarrow$63.6} & 0.087$\rightarrow$0.073 \\
& {-v3} & 0.9 & 801$\rightarrow$554.1 & \textbf{137 $\rightarrow$ 40.2} & 0.054$\rightarrow$0.041 \\
\hline
\end{tabular}
 \begin{tablenotes}
        \footnotesize
        \item 
        		We ran each experiment 10 times and computed the average numbers.
  \end{tablenotes}
\end{threeparttable}
\label{tab:efficiency-results}

\end{table}

\begin{figure}[htbp]
\begin{center}
\subfloat[Task 1]{
\includegraphics[width=0.5\columnwidth]{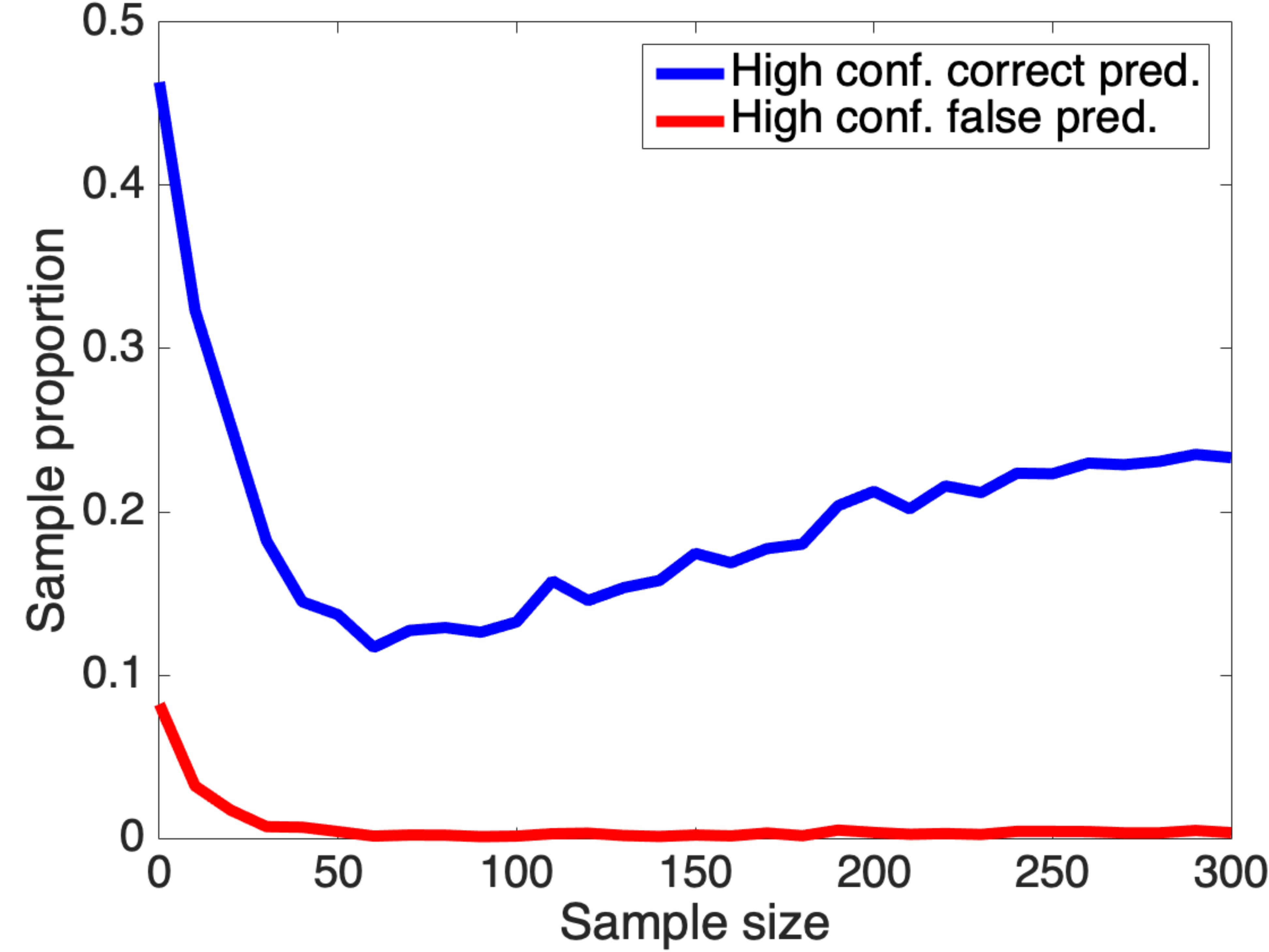}}
\subfloat[Task 2]{
\includegraphics[width=0.5\columnwidth]{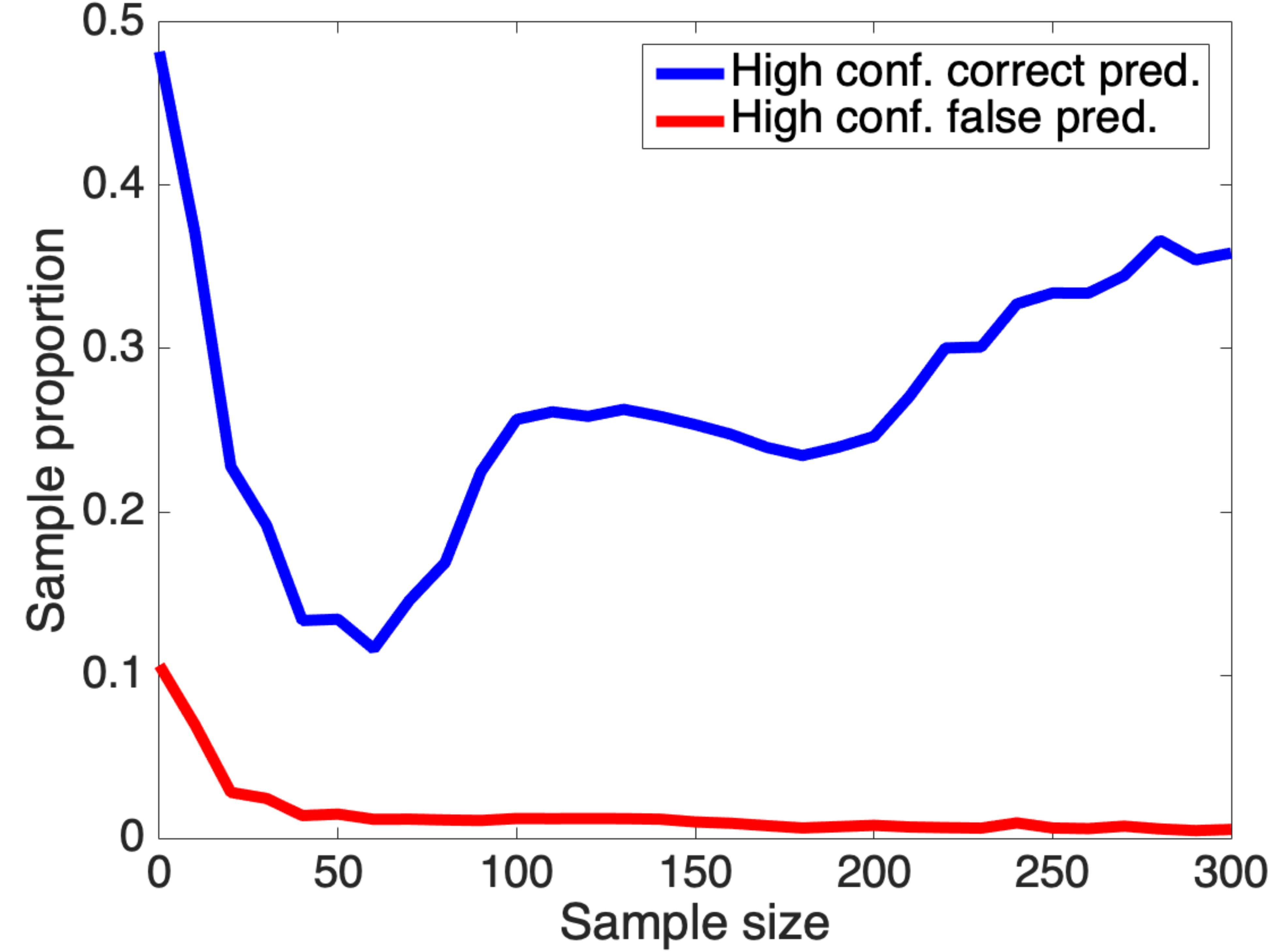}}

\vspace{-2ex}
\subfloat[Task 3]{
\includegraphics[width=0.5\columnwidth]{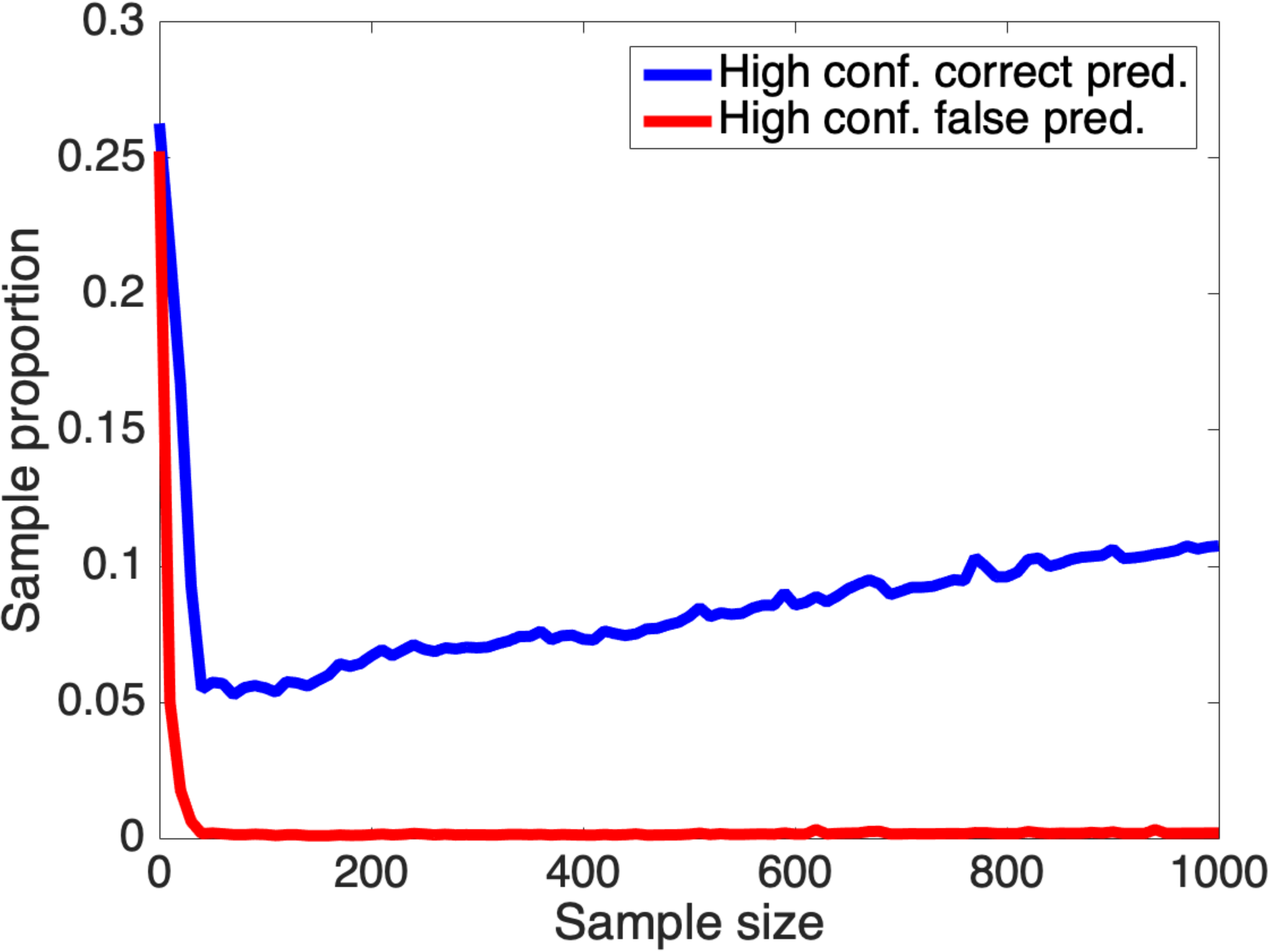}}
\subfloat[Task 4]{
\includegraphics[width=0.5\columnwidth]{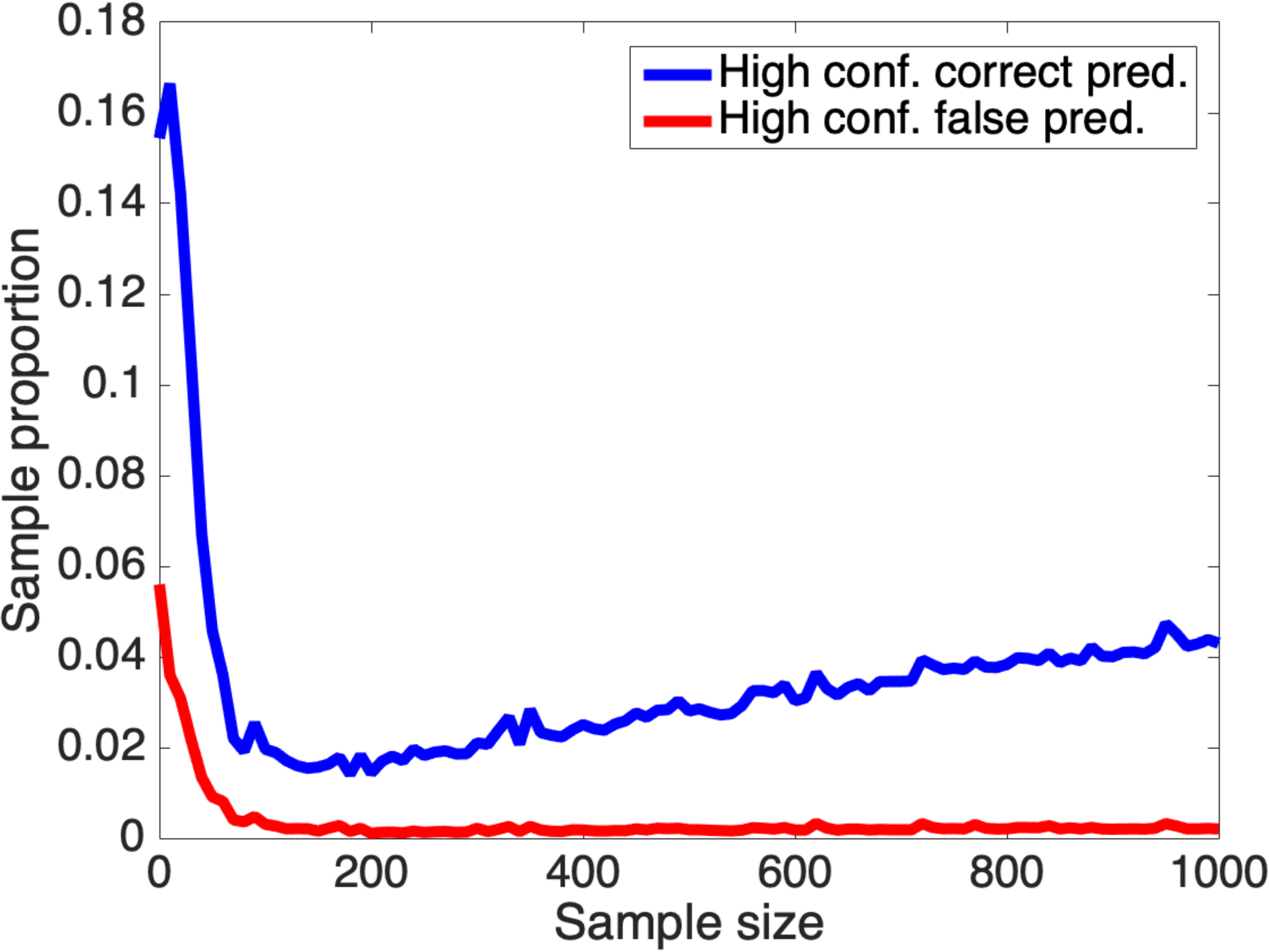}}

\vspace{-2ex}
\subfloat[Task 5]{
\includegraphics[width=0.5\columnwidth]{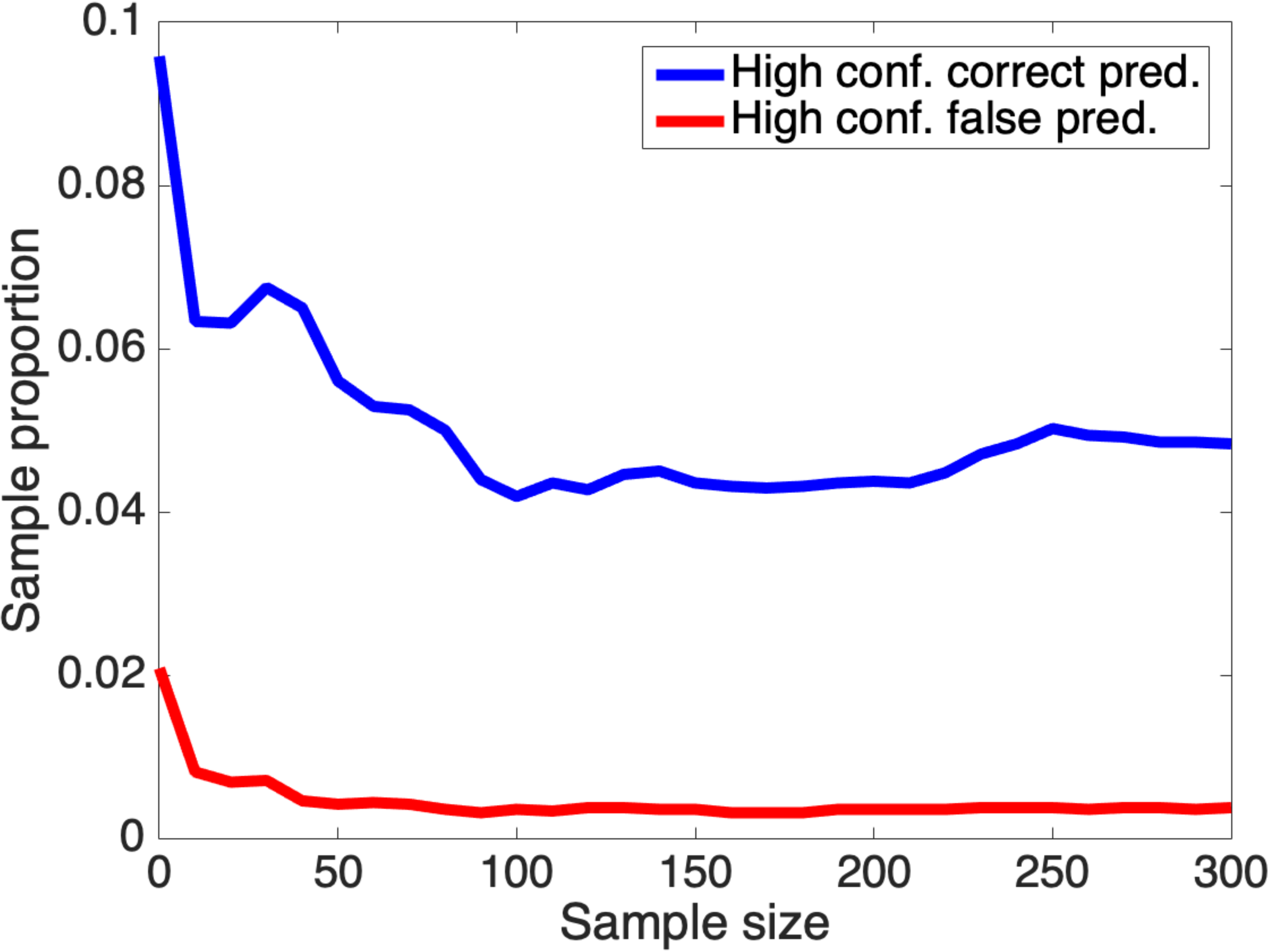}}
\subfloat[Task 6]{
\includegraphics[width=0.5\columnwidth]{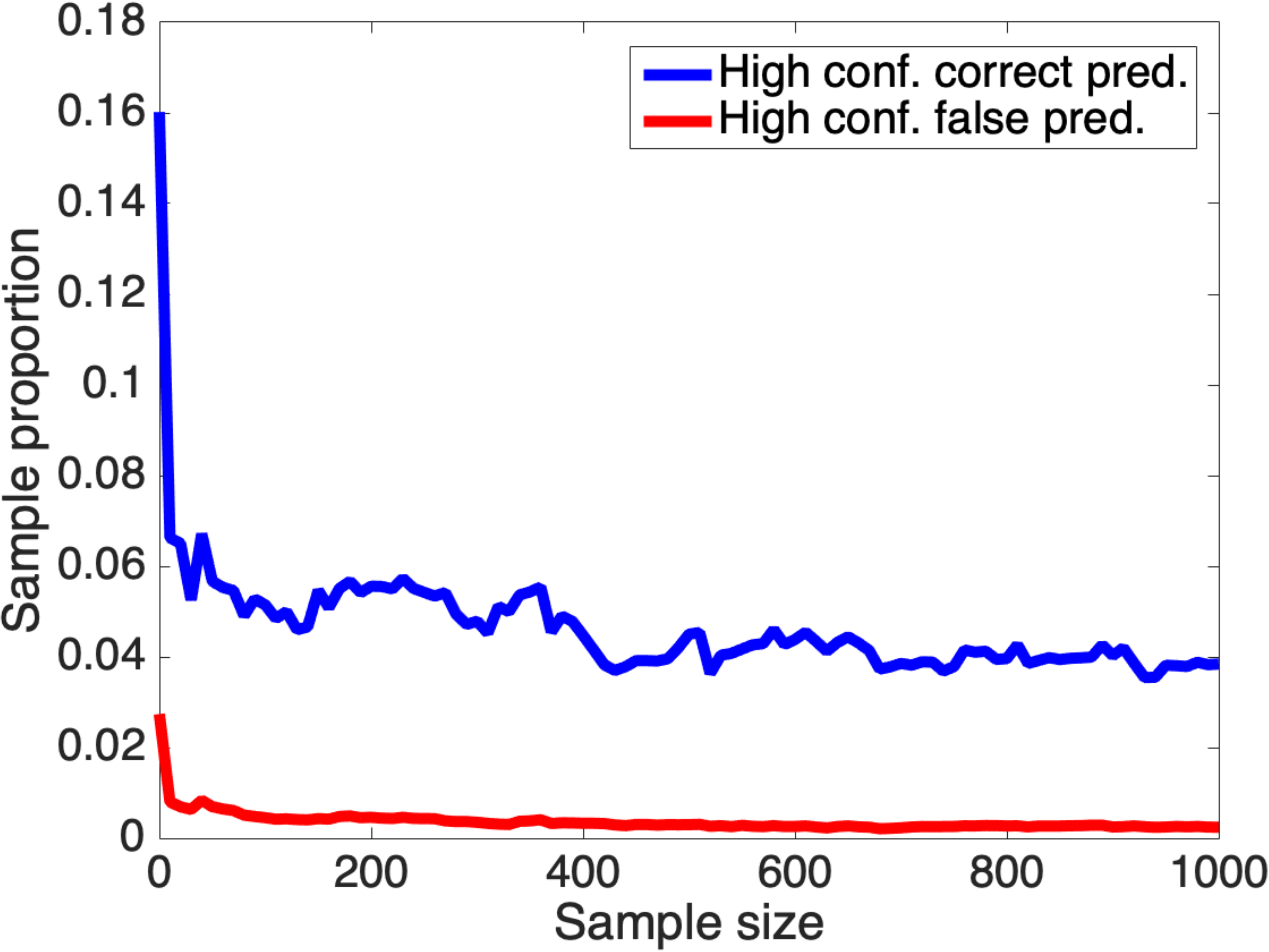}}
\caption{The proportion curve of high confidence inputs.
\added{\rm{Sample size 0 means uncalibrated. The calibration started to take effect with very few data.}}}
\label{fig:efficiency-curve}
\end{center}
\end{figure}

For a visual illustration of the efficiency of our approach, 
Figure~\ref{fig:efficiency-curve}  plots the change of proportions of 
high-confidence false and correct predictions as the size of data used in calibration increases. 
It is interesting to see that: 
(1) most of the high-confidence false predictions were identified very quickly, and
(2) the approach was conservative, 
but the conservativeness is gradually remedied with more labeled operation data used. 

Note that for tasks 4, 5 and 6, 
usual fine-tuning tricks did not work even with all the operation data labeled.
With our operational calibration, using only about 10\% of the data, 
we avoided about 97\%, 80\%, and 71\% high-confidence ($>$$0.9$) errors, respectively.

Based on the results, we can say that \emph{the Gaussian Process-based operational calibration 
is efficient in detecting most of the high-confidence errors with a small amount of labeled operation data}.

\section{Related work}
\label{sec:relatedwork}
Operational calibration is 
generally related to the \added{quality assurance for} \deleted{testing of} deep learning systems 
in the software engineering community, and the confidence calibration, 
transfer learning, and active learning in the machine learning community.   
We briefly overview related work in these directions 
and highlight the connections and differences between our work and them. 

\subsection{Software \added{quality assurance} \deleted{testing} for deep learning systems}
The research in this area can be roughly classified into four categories according to 
the kind of defects targeted: 
\begin{itemize}[leftmargin=*]
\item \emph{Defects in DL programs}. This line of work focuses on 
the bugs in the code of DL frameworks. 
For example, \citeauthor{pham2019cradle} proposed to test the implementation 
of deep learning libraries (TensorFlow, CNTK and Theano) through differential testing~\cite{pham2019cradle}.
\citeauthor{odena2019tensorfuzz} used fuzzing techniques to expose 
numerical errors in matrix multiplication operations~\cite{odena2019tensorfuzz}. 
\item \emph{Defects in DL models}. 
Regarding trained DNN models as pieces of software artifact, 
and borrowing the idea of structural coverage in conventional software testing, 
a series of coverage criteria have been proposed for the testing of DNNs, 
for example, 
%
DeepXplore~\cite{pei2017deepxplore}, DeepGauge~\cite{ma2018deepgauge}, 
DeepConcolic~\cite{sun2018concolic}, and Surprise Adequacy~\cite{kim2019guiding},
to name but a few. 


\item \emph{Defects in training  datasets}.
Another critical element in machine learning is the dataset. 
There exist research aiming at debugging and fixing errors in the polluted training dataset. 
For example, PSI identifies root causes (e.g., incorrect labels) of data errors 
by efficiently computing the Probability of Sufficiency scores through 
probabilistic programming~\cite{chakarov2016debugging}.
\item \emph{Defects due to improper inputs}. A DNN model cannot well handle inputs 
out of the distribution for which it is trained. 
Thus a defensive approach is to detect such inputs. 
For example, \citeauthor{wang2019adversarial}'s approach checked whether an input is normal or adversarial 
by integrating statistical hypothesis testing and model mutation testing~\cite{wang2019adversarial}. 
\citeauthor{wang2020DISSECTOR} proposed DISSECTOR, which effectively distinguished unexpected inputs from normal inputs by verifying progressive relationship between layers~\cite{wang2020DISSECTOR}.
More work in this line can be found in the machine learning literature 
under the name of out-of-distribution detection~\cite{shalev2018out}.
\end{itemize}
For a more comprehensive survey on the testing of machine learning systems, one can consult 
 \citeauthor{zhang2019surveyMLT}~\cite{zhang2019surveyMLT}.

The main difference of our work, compared with these pieces of research, 
is that it is \emph{operational}, i.e., focusing on how well a DNN model will work 
in a given operation domain. As discussed in Section~\ref{sec:operqa}, 
without considering the operation domain, 
it is often difficult to tell whether a phenomena of a DNN model is a bug or a feature~\cite{ilyas2019adversarial,li2019structural}.

An exception is the recent proposal of \emph{operational testing}  
 for the efficient estimation of the accuracy of a DNN model 
in the field~\cite{li2019boosting}. 
Arguably operational calibration is more challenging and more rewarding than 
operational testing, because the latter only tells the overall performance 
of a model in an operation domain, but the former tells when it works well and when not. 
 

\subsection{DNN confidence calibration}
Confidence calibration is important for training high quality classifiers. 
There is a plethora of proposals on this topic in the machine learning literature~\cite{niculescu2005predicting,naeini2015obtaining,zadrozny2002transforming, flach2016calibration, guo2017calibration}. Apart from the Temperature Scaling discussed in 
Section~\ref{subsec:dnnconfidence}, 
Isotonic regression~\cite{zadrozny2002transforming}, Histogram binning~\cite{zadrozny2001obtaining}, and Platt scaling~\cite{Platt1999probabilisticoutputs} 
are also often used. 
Isotonic regression is a non-parametric approach that employs the least square method with a non-decreasing and piecewise constant fitted function.
Histogram binning divides confidences into mutually exclusive bins 
and assigns the calibrated confidences by minimizing the bin-wise squared loss.
Platt scaling is a generalized version of Temperature Scaling.
It adds a linear transformation between the logit layer and the softmax layer, 
and optimizes the parameters with the NLL loss.
However, according to \citeauthor{guo2017calibration}, 
Temperature Scaling is often the most effective approach~\cite{guo2017calibration}.  

As discussed earlier in Section~\ref{subsec:discussions},
the problem of these calibration methods is that they regard 
confidence errors as systematic errors, 
which is usually not the case in the operation domain. 
Technically, these calibration methods are effective in 
minimize the reliability part of the Brier score, 
but ineffective in dealing with the problem in the resolution part. 

In addition, \citeauthor{flach2016calibration} discussed the problem of confidence calibration 
from a decision-theoretic perspective~\cite{flach2016calibration}. 
However, the confidence error caused by domain shift was not explicitly addressed. 


%

Previous research efforts on confidence calibration at prediction time instead of training time are uncommon, but do exist. 
\citeauthor{gal2016dropout} proposed to build a temporary ensemble model by using dropout at prediction time~\cite{gal2016dropout}. 
Despite its elegant Bayesian inference framework, the method is computationally too expensive to handle 
large-scale tasks. 
Recently, based on the insight of conformal prediction~\cite{shafer2008tutorial}, \citeauthor{papernot2018deep} 
proposed to build a k-Nearest Neighbors (kNN) model for the output of each DNN immediate layer~\cite{papernot2018deep}.   
The confidence of a prediction was estimated by the conformity of these kNNs' outputs.
Unfortunately, this method is  not applicable to non-trivial tasks either because of its incapability in handling high-dimensional examples. 
Note that these methods did not explicitly consider confidence errors caused by domain shifts.

\added{Another related line of work is under the name of uncertainty estimation~\cite{sensoy2018evidential, kuleshov2018accurate, schulam2019can}. 
For example, Evidential Deep Learning qualitatively evaluates the uncertainty of DNN predictions 
with a Dirichlet distribution placed on the class probabilities~\cite{sensoy2018evidential}. 
However, these methods mainly aim at out-of-distribution detection, which is considered easier than confidence calibration~\cite{flach2016calibration}. 
Identifying out-of-distribution inputs is useful in defending against adversarial attacks, 
but not directly helpful in adapting a model to a new operation domain. 
}

\subsection{Transfer learning and active learning}
Our approach to operational calibration borrowed ideas from 
transfer learning~\cite{pan2009survey}
and active learning~\cite{settles2009active}. 
Transfer learning (or domain adaptation) aims at training a model from a source domain (origin domain in our terms)
that can be generalized to a target domain (operation domain), 
despite the dataset shift~\cite{ng2016nuts} between the domains. 
The key is to learn features that are transferable between the domains. 

However, transfer learning techniques usually require 
data from both of the source and target domains. 
Contrastingly, operational calibration often has to work 
with limited data from the operation domain and no data from the origin domain. 
Transfer learning usually fails to work under this constraint. 
Even in case it works, it does not necessarily produce well calibrated models, 
and operational calibration is  needed to correct confidence errors~(cf. Figure~\ref{fig:ft-cali3}). 

Active learning aims at reducing the cost of labeling training data by 
deliberately selecting and labeling inputs from a large set of unlabeled data.  
For the Gaussian Process Regression, 
there exist different input selection strategies~\cite{seo2000gaussian, kapoor2007active, pasolli2011gaussian}.
We tried many of them, such as those based on uncertainty~\cite{seo2000gaussian}, on density~\cite{zhu2009active}, 
and on disagreement~\cite{pasolli2011gaussian}, but failed to find a reliable strategy 
that can further improve the data efficiency of our approach.
They were very sensitive to the choices of the initial inputs, the models, 
and the distribution of examples~\cite{settles2009active}.
However, we found that the combination of cost-sensitive sampling bias and uncertainty 
can help in reducing high-confidence false predictions, especially in a cost-sensitive setting.

%

\section{conclusion}
\label{sec:conclusion}

Software quality assurance for systems incorporating DNN models 
is urgently needed. 
This paper focuses on the problem of operational calibration that 
detects and fixes the errors in the confidence given by a DNN model 
for its predictions in a given operation domain.
A Bayesian approach to operational calibration 
is given. It solves the problem with Gaussian Process Regression, 
which leverages the locality of the operation data, 
and also of their prediction correctness, in the representation space. 
\replaced{Experiments with representative datasets and DNN models
confirmed that the approach can significantly reduce the risk of 
high-confidence false prediction with a small number of  labeled data,
and thus efficiently improve the models' quality of service in operational settings.
}{The approach achieved impressive efficacy and efficiency 
in experiments with representative datasets and DNN models.}

While with empirical evidence, we consider conducting more theoretical analysis on aspects
such as the data efficiency and 
the convergence of our algorithm as future work. 
In addition, we plan to investigate operational calibration methods 
for real-world decisions with more complicated cost models.

\begin{acks}
   We thank the anonymous reviewers for their suggestions. 
   This work is supported by the National Natural Science Foundation of China (61690204, 61932021, 61802170) and the Collaborative Innovation Center of Novel Software Technology and Industrialization.
\end{acks}


\bibliographystyle{ACM-Reference-Format}
\bibliography{main}


\begin{thebibliography}{72}


\ifx \showCODEN    \undefined \def \showCODEN     #1{\unskip}     \fi
\ifx \showDOI      \undefined \def \showDOI       #1{#1}\fi
\ifx \showISBNx    \undefined \def \showISBNx     #1{\unskip}     \fi
\ifx \showISBNxiii \undefined \def \showISBNxiii  #1{\unskip}     \fi
\ifx \showISSN     \undefined \def \showISSN      #1{\unskip}     \fi
\ifx \showLCCN     \undefined \def \showLCCN      #1{\unskip}     \fi
\ifx \shownote     \undefined \def \shownote      #1{#1}          \fi
\ifx \showarticletitle \undefined \def \showarticletitle #1{#1}   \fi
\ifx \showURL      \undefined \def \showURL       {\relax}        \fi
\providecommand\bibfield[2]{#2}
\providecommand\bibinfo[2]{#2}
\providecommand\natexlab[1]{#1}
\providecommand\showeprint[2][]{arXiv:#2}

\bibitem[\protect\citeauthoryear{Bengio, Courville, and Vincent}{Bengio
  et~al\mbox{.}}{2012}]%
        {bengio2012unsupervised}
\bibfield{author}{\bibinfo{person}{Yoshua Bengio}, \bibinfo{person}{Aaron~C
  Courville}, {and} \bibinfo{person}{Pascal Vincent}.}
  \bibinfo{year}{2012}\natexlab{}.
\newblock \showarticletitle{Unsupervised feature learning and deep learning: A
  review and new perspectives}.
\newblock \bibinfo{journal}{\emph{CoRR, abs/1206.5538}}  \bibinfo{volume}{1}
  (\bibinfo{year}{2012}), \bibinfo{pages}{2012}.
\newblock


\bibitem[\protect\citeauthoryear{Bishop}{Bishop}{2006}]%
        {bishop2006prml}
\bibfield{author}{\bibinfo{person}{Christopher~M Bishop}.}
  \bibinfo{year}{2006}\natexlab{}.
\newblock \bibinfo{booktitle}{\emph{{Pattern recognition and machine
  learning}}}.
\newblock \bibinfo{publisher}{Springer}, \bibinfo{address}{New York, NY}.
\newblock
\urldef\tempurl%
\url{http://cds.cern.ch/record/998831}
\showURL{%
\tempurl}
\newblock
\shownote{Softcover published in 2016.}


\bibitem[\protect\citeauthoryear{Bojarski, Testa, Dworakowski, Firner, Flepp,
  Goyal, Jackel, Monfort, Muller, Zhang, Zhang, Zhao, and Zieba}{Bojarski
  et~al\mbox{.}}{2016}]%
        {bojarski2016selfdriven}
\bibfield{author}{\bibinfo{person}{Mariusz Bojarski},
  \bibinfo{person}{Davide~Del Testa}, \bibinfo{person}{Daniel Dworakowski},
  \bibinfo{person}{Bernhard Firner}, \bibinfo{person}{Beat Flepp},
  \bibinfo{person}{Prasoon Goyal}, \bibinfo{person}{Lawrence~D. Jackel},
  \bibinfo{person}{Mathew Monfort}, \bibinfo{person}{Urs Muller},
  \bibinfo{person}{Jiakai Zhang}, \bibinfo{person}{Xin Zhang},
  \bibinfo{person}{Jake Zhao}, {and} \bibinfo{person}{Karol Zieba}.}
  \bibinfo{year}{2016}\natexlab{}.
\newblock \showarticletitle{End to End Learning for Self-Driving Cars}.
\newblock \bibinfo{journal}{\emph{CoRR}}  \bibinfo{volume}{abs/1604.07316}
  (\bibinfo{year}{2016}), 9.
\newblock
\showeprint[arxiv]{1604.07316}
\urldef\tempurl%
\url{http://arxiv.org/abs/1604.07316}
\showURL{%
\tempurl}


\bibitem[\protect\citeauthoryear{Brier}{Brier}{1950}]%
        {brier1950verification}
\bibfield{author}{\bibinfo{person}{Glenn~W Brier}.}
  \bibinfo{year}{1950}\natexlab{}.
\newblock \showarticletitle{Verification of forecasts expressed in terms of
  probability}.
\newblock \bibinfo{journal}{\emph{Monthly weather review}}
  \bibinfo{volume}{78}, \bibinfo{number}{1} (\bibinfo{year}{1950}),
  \bibinfo{pages}{1--3}.
\newblock


\bibitem[\protect\citeauthoryear{Burkardt}{Burkardt}{2014}]%
        {burkardt2014truncated}
\bibfield{author}{\bibinfo{person}{John Burkardt}.}
  \bibinfo{year}{2014}\natexlab{}.
\newblock \bibinfo{title}{The truncated normal distribution}.
\newblock , \bibinfo{numpages}{32}~pages.
\newblock
\urldef\tempurl%
\url{http://people.sc.fsu.edu/\~jburkardt/presentations/truncated normal.pdf}
\showURL{%
\tempurl}


\bibitem[\protect\citeauthoryear{Chakarov, Nori, Rajamani, Sen, and
  Vijaykeerthy}{Chakarov et~al\mbox{.}}{2016}]%
        {chakarov2016debugging}
\bibfield{author}{\bibinfo{person}{Aleksandar Chakarov},
  \bibinfo{person}{Aditya Nori}, \bibinfo{person}{Sriram Rajamani},
  \bibinfo{person}{Shayak Sen}, {and} \bibinfo{person}{Deepak Vijaykeerthy}.}
  \bibinfo{year}{2016}\natexlab{}.
\newblock \showarticletitle{Debugging machine learning tasks}.
\newblock \bibinfo{journal}{\emph{arXiv preprint arXiv:1603.07292}}
  (\bibinfo{year}{2016}), 23.
\newblock


\bibitem[\protect\citeauthoryear{Coates, Ng, and Lee}{Coates
  et~al\mbox{.}}{2011}]%
        {coates2011analysis}
\bibfield{author}{\bibinfo{person}{Adam Coates}, \bibinfo{person}{Andrew Ng},
  {and} \bibinfo{person}{Honglak Lee}.} \bibinfo{year}{2011}\natexlab{}.
\newblock \showarticletitle{An analysis of single-layer networks in
  unsupervised feature learning}. In \bibinfo{booktitle}{\emph{Proceedings of
  the fourteenth international conference on artificial intelligence and
  statistics}}. \bibinfo{publisher}{aistats}, \bibinfo{pages}{215--223}.
\newblock


\bibitem[\protect\citeauthoryear{{Council of European Union}}{{Council of
  European Union}}{2014}]%
        {eu-269-2014}
\bibfield{author}{\bibinfo{person}{{Council of European Union}}.}
  \bibinfo{year}{2014}\natexlab{}.
\newblock \bibinfo{title}{Council regulation ({EU}) no 269/2014}.
\newblock
\newblock
\newblock
\shownote{\newline\url{http://eur-lex.europa.eu/legal-content/EN/TXT/?qid=1416170084502&uri=CELEX:32014R0269}.}


\bibitem[\protect\citeauthoryear{Deng, Dong, Socher, jia Li, Li, and
  Fei-fei}{Deng et~al\mbox{.}}{2009}]%
        {Deng09imagenet}
\bibfield{author}{\bibinfo{person}{Jia Deng}, \bibinfo{person}{Wei Dong},
  \bibinfo{person}{Richard Socher}, \bibinfo{person}{Li jia Li},
  \bibinfo{person}{Kai Li}, {and} \bibinfo{person}{Li Fei-fei}.}
  \bibinfo{year}{2009}\natexlab{}.
\newblock \showarticletitle{Imagenet: A large-scale hierarchical image
  database}. In \bibinfo{booktitle}{\emph{In CVPR}}. \bibinfo{publisher}{CVPR},
  8.
\newblock


\bibitem[\protect\citeauthoryear{Flach}{Flach}{2016}]%
        {flach2016calibration}
\bibfield{author}{\bibinfo{person}{Peter~A. Flach}.}
  \bibinfo{year}{2016}\natexlab{}.
\newblock \bibinfo{booktitle}{\emph{Classifier Calibration}}.
\newblock \bibinfo{publisher}{Springer US}, \bibinfo{address}{Boston, MA},
  \bibinfo{pages}{1--8}.
\newblock
\showISBNx{978-1-4899-7502-7}
\urldef\tempurl%
\url{https://doi.org/10.1007/978-1-4899-7502-7_900-1}
\showDOI{\tempurl}


\bibitem[\protect\citeauthoryear{Friedman, Hastie, and Tibshirani}{Friedman
  et~al\mbox{.}}{2001}]%
        {friedman2001elements}
\bibfield{author}{\bibinfo{person}{Jerome Friedman}, \bibinfo{person}{Trevor
  Hastie}, {and} \bibinfo{person}{Robert Tibshirani}.}
  \bibinfo{year}{2001}\natexlab{}.
\newblock \bibinfo{booktitle}{\emph{The elements of statistical learning}}.
  Vol.~\bibinfo{volume}{1}.
\newblock \bibinfo{publisher}{Springer series in statistics New York}.
\newblock


\bibitem[\protect\citeauthoryear{Gal and Ghahramani}{Gal and
  Ghahramani}{2015}]%
        {gal2016dropout}
\bibfield{author}{\bibinfo{person}{Yarin Gal} {and} \bibinfo{person}{Zoubin
  Ghahramani}.} \bibinfo{year}{2015}\natexlab{}.
\newblock \showarticletitle{Dropout as a bayesian approximation: Representing
  model uncertainty in deep learning}.
\newblock \bibinfo{journal}{\emph{arXiv preprint arXiv:1506.02142}}
  (\bibinfo{year}{2015}), 12.
\newblock


\bibitem[\protect\citeauthoryear{Goodfellow, Bengio, and Courville}{Goodfellow
  et~al\mbox{.}}{2016}]%
        {Goodfellow-et-al-2016}
\bibfield{author}{\bibinfo{person}{Ian Goodfellow}, \bibinfo{person}{Yoshua
  Bengio}, {and} \bibinfo{person}{Aaron Courville}.}
  \bibinfo{year}{2016}\natexlab{}.
\newblock \bibinfo{booktitle}{\emph{Deep Learning}}.
\newblock \bibinfo{publisher}{MIT Press}, \bibinfo{address}{New York, NY, USA}.
\newblock
\newblock
\shownote{\url{http://www.deeplearningbook.org}.}


\bibitem[\protect\citeauthoryear{Guo, Pleiss, Sun, and Weinberger}{Guo
  et~al\mbox{.}}{2017}]%
        {guo2017calibration}
\bibfield{author}{\bibinfo{person}{Chuan Guo}, \bibinfo{person}{Geoff Pleiss},
  \bibinfo{person}{Yu Sun}, {and} \bibinfo{person}{Kilian~Q. Weinberger}.}
  \bibinfo{year}{2017}\natexlab{}.
\newblock \showarticletitle{On Calibration of Modern Neural Networks}. In
  \bibinfo{booktitle}{\emph{Proceedings of the 34th International Conference on
  Machine Learning - Volume 70}} (Sydney, NSW, Australia)
  \emph{(\bibinfo{series}{ICML'17})}. \bibinfo{publisher}{JMLR.org},
  \bibinfo{pages}{1321--1330}.
\newblock
\urldef\tempurl%
\url{http://dl.acm.org/citation.cfm?id=3305381.3305518}
\showURL{%
\tempurl}


\bibitem[\protect\citeauthoryear{Hastie, Tibshirani, and Friedman}{Hastie
  et~al\mbox{.}}{2009}]%
        {hastie2009elements}
\bibfield{author}{\bibinfo{person}{T. Hastie}, \bibinfo{person}{R. Tibshirani},
  {and} \bibinfo{person}{J.H. Friedman}.} \bibinfo{year}{2009}\natexlab{}.
\newblock \bibinfo{booktitle}{\emph{The Elements of Statistical Learning: Data
  Mining, Inference, and Prediction}}.
\newblock \bibinfo{publisher}{Springer}.
\newblock
\showISBNx{9780387848846}
\showLCCN{2008941148}
\urldef\tempurl%
\url{https://books.google.com/books?id=eBSgoAEACAAJ}
\showURL{%
\tempurl}


\bibitem[\protect\citeauthoryear{Hinton, Vinyals, and Dean}{Hinton
  et~al\mbox{.}}{2015}]%
        {hinton2015distilling}
\bibfield{author}{\bibinfo{person}{Geoffrey Hinton}, \bibinfo{person}{Oriol
  Vinyals}, {and} \bibinfo{person}{Jeff Dean}.}
  \bibinfo{year}{2015}\natexlab{}.
\newblock \showarticletitle{Distilling the knowledge in a neural network}.
\newblock \bibinfo{journal}{\emph{arXiv preprint arXiv:1503.02531}}
  (\bibinfo{year}{2015}), 9.
\newblock


\bibitem[\protect\citeauthoryear{Ilyas, Santurkar, Tsipras, Engstrom, Tran, and
  Madry}{Ilyas et~al\mbox{.}}{2175}]%
        {ilyas2019adversarial}
\bibfield{author}{\bibinfo{person}{Andrew Ilyas}, \bibinfo{person}{Shibani
  Santurkar}, \bibinfo{person}{Dimitris Tsipras}, \bibinfo{person}{Logan
  Engstrom}, \bibinfo{person}{Brandon Tran}, {and} \bibinfo{person}{Aleksander
  Madry}.} \bibinfo{year}{abs/1905.02175}\natexlab{}.
\newblock \showarticletitle{Adversarial examples are not bugs, they are
  features}.
\newblock \bibinfo{journal}{\emph{arXiv preprint arXiv:1905.02175}}
  \bibinfo{volume}{0}, \bibinfo{number}{0} (\bibinfo{year}{abs/1905.02175}),
  \bibinfo{pages}{0}.
\newblock


\bibitem[\protect\citeauthoryear{Kapoor, Grauman, Urtasun, and Darrell}{Kapoor
  et~al\mbox{.}}{2007}]%
        {kapoor2007active}
\bibfield{author}{\bibinfo{person}{Ashish Kapoor}, \bibinfo{person}{Kristen
  Grauman}, \bibinfo{person}{Raquel Urtasun}, {and} \bibinfo{person}{Trevor
  Darrell}.} \bibinfo{year}{2007}\natexlab{}.
\newblock \showarticletitle{Active learning with gaussian processes for object
  categorization}. In \bibinfo{booktitle}{\emph{2007 IEEE 11th International
  Conference on Computer Vision}}. IEEE, \bibinfo{publisher}{IEEE},
  \bibinfo{pages}{1--8}.
\newblock


\bibitem[\protect\citeauthoryear{Kim, Feldt, and Yoo}{Kim
  et~al\mbox{.}}{2019}]%
        {kim2019guiding}
\bibfield{author}{\bibinfo{person}{Jinhan Kim}, \bibinfo{person}{Robert Feldt},
  {and} \bibinfo{person}{Shin Yoo}.} \bibinfo{year}{2019}\natexlab{}.
\newblock \showarticletitle{Guiding Deep Learning System Testing Using Surprise
  Adequacy}. In \bibinfo{booktitle}{\emph{Proceedings of the 41st International
  Conference on Software Engineering}} (Montreal, Quebec, Canada)
  \emph{(\bibinfo{series}{ICSE '19})}. \bibinfo{publisher}{IEEE Press},
  \bibinfo{address}{Piscataway, NJ, USA}, \bibinfo{pages}{1039--1049}.
\newblock
\urldef\tempurl%
\url{https://doi.org/10.1109/ICSE.2019.00108}
\showDOI{\tempurl}


\bibitem[\protect\citeauthoryear{Kone{\v{c}}n{\`y}, McMahan, Yu, Richt{\'a}rik,
  Suresh, and Bacon}{Kone{\v{c}}n{\`y} et~al\mbox{.}}{2016}]%
        {konevcny2016federated}
\bibfield{author}{\bibinfo{person}{Jakub Kone{\v{c}}n{\`y}},
  \bibinfo{person}{H~Brendan McMahan}, \bibinfo{person}{Felix~X Yu},
  \bibinfo{person}{Peter Richt{\'a}rik}, \bibinfo{person}{Ananda~Theertha
  Suresh}, {and} \bibinfo{person}{Dave Bacon}.}
  \bibinfo{year}{2016}\natexlab{}.
\newblock \showarticletitle{Federated learning: Strategies for improving
  communication efficiency}.
\newblock \bibinfo{journal}{\emph{arXiv preprint arXiv:1610.05492}}
  \bibinfo{number}{10} (\bibinfo{year}{2016}).
\newblock


\bibitem[\protect\citeauthoryear{Krizhevsky, Hinton, et~al\mbox{.}}{Krizhevsky
  et~al\mbox{.}}{2009}]%
        {krizhevsky2009learning}
\bibfield{author}{\bibinfo{person}{Alex Krizhevsky}, \bibinfo{person}{Geoffrey
  Hinton}, {et~al\mbox{.}}} \bibinfo{year}{2009}\natexlab{}.
\newblock \bibinfo{booktitle}{\emph{Learning multiple layers of features from
  tiny images}}.
\newblock \bibinfo{type}{{T}echnical {R}eport}.
  \bibinfo{institution}{Citeseer}.
\newblock


\bibitem[\protect\citeauthoryear{Kuleshov, Fenner, and Ermon}{Kuleshov
  et~al\mbox{.}}{2018}]%
        {kuleshov2018accurate}
\bibfield{author}{\bibinfo{person}{Volodymyr Kuleshov}, \bibinfo{person}{Nathan
  Fenner}, {and} \bibinfo{person}{Stefano Ermon}.}
  \bibinfo{year}{2018}\natexlab{}.
\newblock \showarticletitle{Accurate uncertainties for deep learning using
  calibrated regression}.
\newblock \bibinfo{journal}{\emph{arXiv preprint arXiv:1807.00263}}
  (\bibinfo{year}{2018}), 9.
\newblock


\bibitem[\protect\citeauthoryear{LeCun, Bengio, and Hinton}{LeCun
  et~al\mbox{.}}{2015}]%
        {LeCun:2015aa}
\bibfield{author}{\bibinfo{person}{Yann LeCun}, \bibinfo{person}{Yoshua
  Bengio}, {and} \bibinfo{person}{Geoffrey Hinton}.}
  \bibinfo{year}{2015}\natexlab{}.
\newblock \showarticletitle{Deep learning}.
\newblock \bibinfo{journal}{\emph{Nature}}  \bibinfo{volume}{521}
  (\bibinfo{date}{27 05} \bibinfo{year}{2015}), \bibinfo{pages}{436 EP --}.
\newblock
\urldef\tempurl%
\url{https://doi.org/10.1038/nature14539}
\showURL{%
\tempurl}


\bibitem[\protect\citeauthoryear{LeCun, Bottou, Bengio, Haffner,
  et~al\mbox{.}}{LeCun et~al\mbox{.}}{1998}]%
        {lecun1998gradient}
\bibfield{author}{\bibinfo{person}{Yann LeCun}, \bibinfo{person}{L{\'e}on
  Bottou}, \bibinfo{person}{Yoshua Bengio}, \bibinfo{person}{Patrick Haffner},
  {et~al\mbox{.}}} \bibinfo{year}{1998}\natexlab{}.
\newblock \showarticletitle{Gradient-based learning applied to document
  recognition}.
\newblock \bibinfo{journal}{\emph{Proc. IEEE}} \bibinfo{volume}{86},
  \bibinfo{number}{11} (\bibinfo{year}{1998}), \bibinfo{pages}{2278--2324}.
\newblock


\bibitem[\protect\citeauthoryear{Li, Ma, Xu, and Cao}{Li
  et~al\mbox{.}}{2019a}]%
        {li2019structural}
\bibfield{author}{\bibinfo{person}{Zenan Li}, \bibinfo{person}{Xiaoxing Ma},
  \bibinfo{person}{Chang Xu}, {and} \bibinfo{person}{Chun Cao}.}
  \bibinfo{year}{2019}\natexlab{a}.
\newblock \showarticletitle{Structural Coverage Criteria for Neural Networks
  Could Be Misleading}. In \bibinfo{booktitle}{\emph{Proceedings of the 41st
  International Conference on Software Engineering: New Ideas and Emerging
  Results}} (Montreal, Quebec, Canada) \emph{(\bibinfo{series}{ICSE-NIER
  '19})}. \bibinfo{publisher}{IEEE Press}, \bibinfo{address}{Piscataway, NJ,
  USA}, \bibinfo{pages}{89--92}.
\newblock
\urldef\tempurl%
\url{https://doi.org/10.1109/ICSE-NIER.2019.00031}
\showDOI{\tempurl}


\bibitem[\protect\citeauthoryear{Li, Ma, Xu, Cao, Xu, and Lu}{Li
  et~al\mbox{.}}{2019b}]%
        {li2019boosting}
\bibfield{author}{\bibinfo{person}{Zenan Li}, \bibinfo{person}{Xiaoxing Ma},
  \bibinfo{person}{Chang Xu}, \bibinfo{person}{Chun Cao},
  \bibinfo{person}{Jingwei Xu}, {and} \bibinfo{person}{Jian Lu}.}
  \bibinfo{year}{2019}\natexlab{b}.
\newblock \showarticletitle{Boosting Operational {DNN} Testing Efficiency
  through Conditioning}. In \bibinfo{booktitle}{\emph{Proceedings of the 27th
  ACM Joint European Software Engineering Conference and Symposium on the
  Foundations of Software Engineering}} \emph{(\bibinfo{series}{ESEC/FSE
  '19})}. \bibinfo{publisher}{ACM}, \bibinfo{address}{Tallinn, Estonia}, 12.
\newblock
\urldef\tempurl%
\url{http://arxiv.org/abs/1906.02533}
\showURL{%
\tempurl}


\bibitem[\protect\citeauthoryear{Liu, Gadepalli, Norouzi, Dahl, Kohlberger,
  Boyko, Venugopalan, Timofeev, Nelson, Corrado, Hipp, Peng, and Stumpe}{Liu
  et~al\mbox{.}}{2017}]%
        {liu2017detecting}
\bibfield{author}{\bibinfo{person}{Yun Liu}, \bibinfo{person}{Krishna
  Gadepalli}, \bibinfo{person}{Mohammad Norouzi}, \bibinfo{person}{George~E.
  Dahl}, \bibinfo{person}{Timo Kohlberger}, \bibinfo{person}{Aleksey Boyko},
  \bibinfo{person}{Subhashini Venugopalan}, \bibinfo{person}{Aleksei Timofeev},
  \bibinfo{person}{Philip~Q. Nelson}, \bibinfo{person}{Greg~S. Corrado},
  \bibinfo{person}{Jason~D. Hipp}, \bibinfo{person}{Lily Peng}, {and}
  \bibinfo{person}{Martin~C. Stumpe}.} \bibinfo{year}{2017}\natexlab{}.
\newblock \bibinfo{title}{Detecting Cancer Metastases on Gigapixel Pathology
  Images}.
\newblock
\newblock
\showeprint[arxiv]{1703.02442}~[cs.CV]


\bibitem[\protect\citeauthoryear{{Ma}, {Juefei-Xu}, {Xue}, {Li}, {Li}, {Liu},
  and {Zhao}}{{Ma} et~al\mbox{.}}{2019}]%
        {ma2018combinatorial}
\bibfield{author}{\bibinfo{person}{L. {Ma}}, \bibinfo{person}{F. {Juefei-Xu}},
  \bibinfo{person}{M. {Xue}}, \bibinfo{person}{B. {Li}}, \bibinfo{person}{L.
  {Li}}, \bibinfo{person}{Y. {Liu}}, {and} \bibinfo{person}{J. {Zhao}}.}
  \bibinfo{year}{2019}\natexlab{}.
\newblock \showarticletitle{DeepCT: Tomographic Combinatorial Testing for Deep
  Learning Systems}. In \bibinfo{booktitle}{\emph{2019 IEEE 26th International
  Conference on Software Analysis, Evolution and Reengineering (SANER)}}.
  \bibinfo{publisher}{ACM}, \bibinfo{address}{New York, NY, USA},
  \bibinfo{pages}{614--618}.
\newblock
\showISSN{1534-5351}
\urldef\tempurl%
\url{https://doi.org/10.1109/SANER.2019.8668044}
\showDOI{\tempurl}


\bibitem[\protect\citeauthoryear{Ma, Juefei-Xu, Zhang, Sun, Xue, Li, Chen, Su,
  Li, Liu, Zhao, and Wang}{Ma et~al\mbox{.}}{2018}]%
        {ma2018deepgauge}
\bibfield{author}{\bibinfo{person}{Lei Ma}, \bibinfo{person}{Felix Juefei-Xu},
  \bibinfo{person}{Fuyuan Zhang}, \bibinfo{person}{Jiyuan Sun},
  \bibinfo{person}{Minhui Xue}, \bibinfo{person}{Bo Li},
  \bibinfo{person}{Chunyang Chen}, \bibinfo{person}{Ting Su},
  \bibinfo{person}{Li Li}, \bibinfo{person}{Yang Liu}, \bibinfo{person}{Jianjun
  Zhao}, {and} \bibinfo{person}{Yadong Wang}.} \bibinfo{year}{2018}\natexlab{}.
\newblock \showarticletitle{DeepGauge: Multi-granularity Testing Criteria for
  Deep Learning Systems}. In \bibinfo{booktitle}{\emph{Proceedings of the 33rd
  ACM/IEEE International Conference on Automated Software Engineering}}
  (Montpellier, France) \emph{(\bibinfo{series}{ASE 2018})}.
  \bibinfo{publisher}{ACM}, \bibinfo{address}{New York, NY, USA},
  \bibinfo{pages}{120--131}.
\newblock
\showISBNx{978-1-4503-5937-5}
\urldef\tempurl%
\url{https://doi.org/10.1145/3238147.3238202}
\showDOI{\tempurl}


\bibitem[\protect\citeauthoryear{Mller, Clough, Deselaers, and Caputo}{Mller
  et~al\mbox{.}}{2010}]%
        {Mller2010ImageCLEF}
\bibfield{author}{\bibinfo{person}{Henning Mller}, \bibinfo{person}{Paul
  Clough}, \bibinfo{person}{Thomas Deselaers}, {and} \bibinfo{person}{Barbara
  Caputo}.} \bibinfo{year}{2010}\natexlab{}.
\newblock \bibinfo{booktitle}{\emph{ImageCLEF: Experimental Evaluation in
  Visual Information Retrieval} (\bibinfo{edition}{1st} ed.)}.
\newblock \bibinfo{publisher}{Springer Publishing Company, Incorporated}.
\newblock
\showISBNx{3642151809, 9783642151804}


\bibitem[\protect\citeauthoryear{Montagnon, Cerny, Cadrin-Ch{\^e}nevert,
  Hamilton, Derennes, Ilinca, Vandenbroucke-Menu, Turcotte, Kadoury, and
  Tang}{Montagnon et~al\mbox{.}}{2020}]%
        {MCM20DLRadiology}
\bibfield{author}{\bibinfo{person}{Emmanuel Montagnon}, \bibinfo{person}{Milena
  Cerny}, \bibinfo{person}{Alexandre Cadrin-Ch{\^e}nevert},
  \bibinfo{person}{Vincent Hamilton}, \bibinfo{person}{Thomas Derennes},
  \bibinfo{person}{Andr{\'e} Ilinca}, \bibinfo{person}{Franck
  Vandenbroucke-Menu}, \bibinfo{person}{Simon Turcotte},
  \bibinfo{person}{Samuel Kadoury}, {and} \bibinfo{person}{An Tang}.}
  \bibinfo{year}{2020}\natexlab{}.
\newblock \showarticletitle{Deep learning workflow in radiology: a primer}.
\newblock \bibinfo{journal}{\emph{Insights into Imaging}} \bibinfo{volume}{11},
  \bibinfo{number}{1} (\bibinfo{year}{2020}), \bibinfo{pages}{22}.
\newblock
\showISBNx{1869-4101}
\urldef\tempurl%
\url{https://doi.org/10.1186/s13244-019-0832-5}
\showDOI{\tempurl}


\bibitem[\protect\citeauthoryear{Murphy}{Murphy}{1973}]%
        {murphy1973BSPartition}
\bibfield{author}{\bibinfo{person}{Allan~H. Murphy}.}
  \bibinfo{year}{1973}\natexlab{}.
\newblock \showarticletitle{A New Vector Partition of the Probability Score}.
\newblock \bibinfo{journal}{\emph{Journal of Applied Meteorology}}
  \bibinfo{volume}{12}, \bibinfo{number}{4} (\bibinfo{year}{1973}),
  \bibinfo{pages}{595--600}.
\newblock
\urldef\tempurl%
\url{https://doi.org/10.1175/1520-0450(1973)012<0595:ANVPOT>2.0.CO;2}
\showDOI{\tempurl}


\bibitem[\protect\citeauthoryear{Naeini, Cooper, and Hauskrecht}{Naeini
  et~al\mbox{.}}{2015}]%
        {naeini2015obtaining}
\bibfield{author}{\bibinfo{person}{Mahdi~Pakdaman Naeini},
  \bibinfo{person}{Gregory Cooper}, {and} \bibinfo{person}{Milos Hauskrecht}.}
  \bibinfo{year}{2015}\natexlab{}.
\newblock \showarticletitle{Obtaining well calibrated probabilities using
  bayesian binning}. In \bibinfo{booktitle}{\emph{Twenty-Ninth AAAI Conference
  on Artificial Intelligence}}. \bibinfo{publisher}{AAAI}, 7.
\newblock


\bibitem[\protect\citeauthoryear{Ng}{Ng}{2016}]%
        {ng2016nuts}
\bibfield{author}{\bibinfo{person}{Andrew Ng}.}
  \bibinfo{year}{2016}\natexlab{}.
\newblock \showarticletitle{Nuts and bolts of building AI applications using
  Deep Learning}.
\newblock \bibinfo{journal}{\emph{Neurips-Keynote}} (\bibinfo{year}{2016}), 5.
\newblock


\bibitem[\protect\citeauthoryear{Niculescu-Mizil and Caruana}{Niculescu-Mizil
  and Caruana}{2005}]%
        {niculescu2005predicting}
\bibfield{author}{\bibinfo{person}{Alexandru Niculescu-Mizil} {and}
  \bibinfo{person}{Rich Caruana}.} \bibinfo{year}{2005}\natexlab{}.
\newblock \showarticletitle{Predicting good probabilities with supervised
  learning}. In \bibinfo{booktitle}{\emph{Proceedings of the 22nd international
  conference on Machine learning}}. ACM, \bibinfo{publisher}{JMLR.org},
  \bibinfo{pages}{625--632}.
\newblock


\bibitem[\protect\citeauthoryear{Obermeyer and Emanuel}{Obermeyer and
  Emanuel}{2016}]%
        {obermeyer2016predicting}
\bibfield{author}{\bibinfo{person}{Ziad Obermeyer} {and}
  \bibinfo{person}{Ezekiel~J Emanuel}.} \bibinfo{year}{2016}\natexlab{}.
\newblock \showarticletitle{Predicting the future---big data, machine learning,
  and clinical medicine}.
\newblock \bibinfo{journal}{\emph{The New England journal of medicine}}
  \bibinfo{volume}{375}, \bibinfo{number}{13} (\bibinfo{year}{2016}),
  \bibinfo{pages}{1216}.
\newblock


\bibitem[\protect\citeauthoryear{Odena, Olsson, Andersen, and Goodfellow}{Odena
  et~al\mbox{.}}{2019}]%
        {odena2019tensorfuzz}
\bibfield{author}{\bibinfo{person}{Augustus Odena}, \bibinfo{person}{Catherine
  Olsson}, \bibinfo{person}{David Andersen}, {and} \bibinfo{person}{Ian
  Goodfellow}.} \bibinfo{year}{2019}\natexlab{}.
\newblock \showarticletitle{{T}ensor{F}uzz: Debugging Neural Networks with
  Coverage-Guided Fuzzing}. In \bibinfo{booktitle}{\emph{Proceedings of the
  36th International Conference on Machine Learning}}
  \emph{(\bibinfo{series}{Proceedings of Machine Learning Research},
  Vol.~\bibinfo{volume}{97})}, \bibfield{editor}{\bibinfo{person}{Kamalika
  Chaudhuri} {and} \bibinfo{person}{Ruslan Salakhutdinov}} (Eds.).
  \bibinfo{publisher}{PMLR}, \bibinfo{address}{Long Beach, California, USA},
  \bibinfo{pages}{4901--4911}.
\newblock
\urldef\tempurl%
\url{http://proceedings.mlr.press/v97/odena19a.html}
\showURL{%
\tempurl}


\bibitem[\protect\citeauthoryear{Pan and Yang}{Pan and Yang}{2009}]%
        {pan2009survey}
\bibfield{author}{\bibinfo{person}{Sinno~Jialin Pan} {and}
  \bibinfo{person}{Qiang Yang}.} \bibinfo{year}{2009}\natexlab{}.
\newblock \showarticletitle{A survey on transfer learning}.
\newblock \bibinfo{journal}{\emph{IEEE Transactions on knowledge and data
  engineering}} \bibinfo{volume}{22}, \bibinfo{number}{10}
  (\bibinfo{year}{2009}), \bibinfo{pages}{1345--1359}.
\newblock


\bibitem[\protect\citeauthoryear{Pang, Lee, and Vaithyanathan}{Pang
  et~al\mbox{.}}{2002}]%
        {Pang2002sentiment}
\bibfield{author}{\bibinfo{person}{Bo Pang}, \bibinfo{person}{Lillian Lee},
  {and} \bibinfo{person}{Shivakumar Vaithyanathan}.}
  \bibinfo{year}{2002}\natexlab{}.
\newblock \showarticletitle{Thumbs Up? Sentiment Classification Using Machine
  Learning Techniques}. In \bibinfo{booktitle}{\emph{Proceedings of EMNLP}}.
  \bibinfo{publisher}{EMNLP}, \bibinfo{pages}{79--86}.
\newblock


\bibitem[\protect\citeauthoryear{Papernot and McDaniel}{Papernot and
  McDaniel}{2018}]%
        {papernot2018deep}
\bibfield{author}{\bibinfo{person}{Nicolas Papernot} {and}
  \bibinfo{person}{Patrick McDaniel}.} \bibinfo{year}{2018}\natexlab{}.
\newblock \showarticletitle{Deep k-nearest neighbors: Towards confident,
  interpretable and robust deep learning}.
\newblock \bibinfo{journal}{\emph{arXiv preprint arXiv:1803.04765}}
  (\bibinfo{year}{2018}), 18.
\newblock


\bibitem[\protect\citeauthoryear{Pasolli and Melgani}{Pasolli and
  Melgani}{2011}]%
        {pasolli2011gaussian}
\bibfield{author}{\bibinfo{person}{Edoardo Pasolli} {and}
  \bibinfo{person}{Farid Melgani}.} \bibinfo{year}{2011}\natexlab{}.
\newblock \showarticletitle{Gaussian process regression within an active
  learning scheme}. In \bibinfo{booktitle}{\emph{2011 IEEE International
  Geoscience and Remote Sensing Symposium}}. IEEE, \bibinfo{publisher}{IEEE},
  \bibinfo{pages}{3574--3577}.
\newblock


\bibitem[\protect\citeauthoryear{Pedregosa, Varoquaux, Gramfort, Michel,
  Thirion, Grisel, Blondel, Prettenhofer, Weiss, Dubourg, Vanderplas, Passos,
  Cournapeau, Brucher, Perrot, and Duchesnay}{Pedregosa et~al\mbox{.}}{2011}]%
        {scikit-learn}
\bibfield{author}{\bibinfo{person}{F. Pedregosa}, \bibinfo{person}{G.
  Varoquaux}, \bibinfo{person}{A. Gramfort}, \bibinfo{person}{V. Michel},
  \bibinfo{person}{B. Thirion}, \bibinfo{person}{O. Grisel},
  \bibinfo{person}{M. Blondel}, \bibinfo{person}{P. Prettenhofer},
  \bibinfo{person}{R. Weiss}, \bibinfo{person}{V. Dubourg}, \bibinfo{person}{J.
  Vanderplas}, \bibinfo{person}{A. Passos}, \bibinfo{person}{D. Cournapeau},
  \bibinfo{person}{M. Brucher}, \bibinfo{person}{M. Perrot}, {and}
  \bibinfo{person}{E. Duchesnay}.} \bibinfo{year}{2011}\natexlab{}.
\newblock \showarticletitle{Scikit-learn: Machine Learning in {P}ython}.
\newblock \bibinfo{journal}{\emph{Journal of Machine Learning Research}}
  \bibinfo{volume}{12} (\bibinfo{year}{2011}), \bibinfo{pages}{2825--2830}.
\newblock


\bibitem[\protect\citeauthoryear{Pei, Cao, Yang, and Jana}{Pei
  et~al\mbox{.}}{2017}]%
        {pei2017deepxplore}
\bibfield{author}{\bibinfo{person}{Kexin Pei}, \bibinfo{person}{Yinzhi Cao},
  \bibinfo{person}{Junfeng Yang}, {and} \bibinfo{person}{Suman Jana}.}
  \bibinfo{year}{2017}\natexlab{}.
\newblock \showarticletitle{DeepXplore: Automated Whitebox Testing of Deep
  Learning Systems}. In \bibinfo{booktitle}{\emph{Proceedings of the 26th
  Symposium on Operating Systems Principles}} (Shanghai, China)
  \emph{(\bibinfo{series}{SOSP '17})}. \bibinfo{publisher}{ACM},
  \bibinfo{address}{New York, NY, USA}, \bibinfo{pages}{1--18}.
\newblock
\showISBNx{978-1-4503-5085-3}
\urldef\tempurl%
\url{https://doi.org/10.1145/3132747.3132785}
\showDOI{\tempurl}


\bibitem[\protect\citeauthoryear{Pham, Lutellier, Qi, and Tan}{Pham
  et~al\mbox{.}}{2019}]%
        {pham2019cradle}
\bibfield{author}{\bibinfo{person}{Hung~Viet Pham}, \bibinfo{person}{Thibaud
  Lutellier}, \bibinfo{person}{Weizhen Qi}, {and} \bibinfo{person}{Lin Tan}.}
  \bibinfo{year}{2019}\natexlab{}.
\newblock \showarticletitle{CRADLE: cross-backend validation to detect and
  localize bugs in deep learning libraries}. In
  \bibinfo{booktitle}{\emph{Proceedings of the 41st International Conference on
  Software Engineering}}. IEEE Press, \bibinfo{publisher}{ICSE'19},
  \bibinfo{pages}{1027--1038}.
\newblock


\bibitem[\protect\citeauthoryear{Platt}{Platt}{1999}]%
        {Platt1999probabilisticoutputs}
\bibfield{author}{\bibinfo{person}{John~C. Platt}.}
  \bibinfo{year}{1999}\natexlab{}.
\newblock \showarticletitle{Probabilistic Outputs for Support Vector Machines
  and Comparisons to Regularized Likelihood Methods}. In
  \bibinfo{booktitle}{\emph{ADVANCES IN LARGE MARGIN CLASSIFIERS}}.
  \bibinfo{publisher}{MIT Press}, \bibinfo{pages}{61--74}.
\newblock


\bibitem[\protect\citeauthoryear{Rasmussen and Williams}{Rasmussen and
  Williams}{2005}]%
        {rasmussen2005GPML}
\bibfield{author}{\bibinfo{person}{Carl~Edward Rasmussen} {and}
  \bibinfo{person}{Christopher K.~I. Williams}.}
  \bibinfo{year}{2005}\natexlab{}.
\newblock \bibinfo{booktitle}{\emph{Gaussian Processes for Machine Learning
  (Adaptive Computation and Machine Learning)}}.
\newblock \bibinfo{publisher}{The MIT Press}.
\newblock
\showISBNx{026218253X}


\bibitem[\protect\citeauthoryear{Riley}{Riley}{2019}]%
        {Riley2019nature}
\bibfield{author}{\bibinfo{person}{Patrick Riley}.}
  \bibinfo{year}{2019}\natexlab{}.
\newblock \showarticletitle{Three pitfalls to avoid in machine learning}.
\newblock \bibinfo{journal}{\emph{Nature}} \bibinfo{volume}{572},
  \bibinfo{number}{7767} (\bibinfo{date}{Jul} \bibinfo{year}{2019}),
  \bibinfo{pages}{27--29}.
\newblock
\showISSN{1476-4687}
\urldef\tempurl%
\url{https://doi.org/10.1038/d41586-019-02307-y}
\showDOI{\tempurl}


\bibitem[\protect\citeauthoryear{Schulam and Saria}{Schulam and Saria}{2019}]%
        {schulam2019can}
\bibfield{author}{\bibinfo{person}{Peter Schulam} {and} \bibinfo{person}{Suchi
  Saria}.} \bibinfo{year}{2019}\natexlab{}.
\newblock \showarticletitle{Can you trust this prediction? Auditing pointwise
  reliability after learning}.
\newblock \bibinfo{journal}{\emph{arXiv preprint arXiv:1901.00403}}
  (\bibinfo{year}{2019}), 10.
\newblock


\bibitem[\protect\citeauthoryear{Sensoy, Kaplan, and Kandemir}{Sensoy
  et~al\mbox{.}}{2018}]%
        {sensoy2018evidential}
\bibfield{author}{\bibinfo{person}{Murat Sensoy}, \bibinfo{person}{Lance
  Kaplan}, {and} \bibinfo{person}{Melih Kandemir}.}
  \bibinfo{year}{2018}\natexlab{}.
\newblock \showarticletitle{Evidential deep learning to quantify classification
  uncertainty}. In \bibinfo{booktitle}{\emph{Advances in Neural Information
  Processing Systems}}. \bibinfo{publisher}{NeurIps},
  \bibinfo{pages}{3179--3189}.
\newblock


\bibitem[\protect\citeauthoryear{Seo, Wallat, Graepel, and Obermayer}{Seo
  et~al\mbox{.}}{2000}]%
        {seo2000gaussian}
\bibfield{author}{\bibinfo{person}{Sambu Seo}, \bibinfo{person}{Marko Wallat},
  \bibinfo{person}{Thore Graepel}, {and} \bibinfo{person}{Klaus Obermayer}.}
  \bibinfo{year}{2000}\natexlab{}.
\newblock \showarticletitle{Gaussian process regression: Active data selection
  and test point rejection}.
\newblock In \bibinfo{booktitle}{\emph{Mustererkennung 2000}}.
  \bibinfo{publisher}{Springer}, \bibinfo{pages}{27--34}.
\newblock


\bibitem[\protect\citeauthoryear{Settles}{Settles}{2009}]%
        {settles2009active}
\bibfield{author}{\bibinfo{person}{Burr Settles}.}
  \bibinfo{year}{2009}\natexlab{}.
\newblock \bibinfo{booktitle}{\emph{Active learning literature survey}}.
\newblock \bibinfo{type}{{T}echnical {R}eport}.
  \bibinfo{institution}{University of Wisconsin-Madison Department of Computer
  Sciences}.
\newblock


\bibitem[\protect\citeauthoryear{Shafer and Vovk}{Shafer and Vovk}{2008}]%
        {shafer2008tutorial}
\bibfield{author}{\bibinfo{person}{Glenn Shafer} {and}
  \bibinfo{person}{Vladimir Vovk}.} \bibinfo{year}{2008}\natexlab{}.
\newblock \showarticletitle{A tutorial on conformal prediction}.
\newblock \bibinfo{journal}{\emph{Journal of Machine Learning Research}}
  \bibinfo{volume}{9}, \bibinfo{number}{Mar} (\bibinfo{year}{2008}),
  \bibinfo{pages}{371--421}.
\newblock


\bibitem[\protect\citeauthoryear{Shalev, Adi, and Keshet}{Shalev
  et~al\mbox{.}}{2018}]%
        {shalev2018out}
\bibfield{author}{\bibinfo{person}{Gabi Shalev}, \bibinfo{person}{Yossi Adi},
  {and} \bibinfo{person}{Joseph Keshet}.} \bibinfo{year}{2018}\natexlab{}.
\newblock \showarticletitle{Out-of-distribution detection using multiple
  semantic label representations}. In \bibinfo{booktitle}{\emph{Advances in
  Neural Information Processing Systems}}. \bibinfo{pages}{7375--7385}.
\newblock


\bibitem[\protect\citeauthoryear{Shokri and Shmatikov}{Shokri and
  Shmatikov}{2015}]%
        {Shokri2015CCS}
\bibfield{author}{\bibinfo{person}{Reza Shokri} {and} \bibinfo{person}{Vitaly
  Shmatikov}.} \bibinfo{year}{2015}\natexlab{}.
\newblock \showarticletitle{Privacy-Preserving Deep Learning}. In
  \bibinfo{booktitle}{\emph{Proceedings of the 22Nd ACM SIGSAC Conference on
  Computer and Communications Security}} (Denver, Colorado, USA)
  \emph{(\bibinfo{series}{CCS '15})}. \bibinfo{publisher}{ACM},
  \bibinfo{address}{New York, NY, USA}, \bibinfo{pages}{1310--1321}.
\newblock
\showISBNx{978-1-4503-3832-5}
\urldef\tempurl%
\url{https://doi.org/10.1145/2810103.2813687}
\showDOI{\tempurl}


\bibitem[\protect\citeauthoryear{Shu, Bui, Narui, and Ermon}{Shu
  et~al\mbox{.}}{2018}]%
        {shu2018a}
\bibfield{author}{\bibinfo{person}{Rui Shu}, \bibinfo{person}{Hung Bui},
  \bibinfo{person}{Hirokazu Narui}, {and} \bibinfo{person}{Stefano Ermon}.}
  \bibinfo{year}{2018}\natexlab{}.
\newblock \showarticletitle{A {DIRT}-T Approach to Unsupervised Domain
  Adaptation}. In \bibinfo{booktitle}{\emph{International Conference on
  Learning Representations}}. \bibinfo{publisher}{ICLR}, 19.
\newblock
\urldef\tempurl%
\url{https://openreview.net/forum?id=H1q-TM-AW}
\showURL{%
\tempurl}


\bibitem[\protect\citeauthoryear{Sun, Wu, Ruan, Huang, Kwiatkowska, and
  Kroening}{Sun et~al\mbox{.}}{2018}]%
        {sun2018concolic}
\bibfield{author}{\bibinfo{person}{Youcheng Sun}, \bibinfo{person}{Min Wu},
  \bibinfo{person}{Wenjie Ruan}, \bibinfo{person}{Xiaowei Huang},
  \bibinfo{person}{Marta Kwiatkowska}, {and} \bibinfo{person}{Daniel
  Kroening}.} \bibinfo{year}{2018}\natexlab{}.
\newblock \showarticletitle{Concolic Testing for Deep Neural Networks}. In
  \bibinfo{booktitle}{\emph{Proceedings of the 33rd ACM/IEEE International
  Conference on Automated Software Engineering}} (Montpellier, France)
  \emph{(\bibinfo{series}{ASE 2018})}. \bibinfo{publisher}{ACM},
  \bibinfo{address}{New York, NY, USA}, \bibinfo{pages}{109--119}.
\newblock
\showISBNx{978-1-4503-5937-5}
\urldef\tempurl%
\url{https://doi.org/10.1145/3238147.3238172}
\showDOI{\tempurl}


\bibitem[\protect\citeauthoryear{Szegedy, Vanhoucke, Ioffe, Shlens, and
  Wojna}{Szegedy et~al\mbox{.}}{2016}]%
        {szegedy2016rethinking}
\bibfield{author}{\bibinfo{person}{Christian Szegedy}, \bibinfo{person}{Vincent
  Vanhoucke}, \bibinfo{person}{Sergey Ioffe}, \bibinfo{person}{Jon Shlens},
  {and} \bibinfo{person}{Zbigniew Wojna}.} \bibinfo{year}{2016}\natexlab{}.
\newblock \showarticletitle{Rethinking the inception architecture for computer
  vision}. In \bibinfo{booktitle}{\emph{Proceedings of the IEEE conference on
  computer vision and pattern recognition}}. \bibinfo{publisher}{CVPR},
  \bibinfo{pages}{2818--2826}.
\newblock


\bibitem[\protect\citeauthoryear{Tewari and Bartlett}{Tewari and
  Bartlett}{2007}]%
        {tewari2007consistency}
\bibfield{author}{\bibinfo{person}{Ambuj Tewari} {and} \bibinfo{person}{Peter~L
  Bartlett}.} \bibinfo{year}{2007}\natexlab{}.
\newblock \showarticletitle{On the consistency of multiclass classification
  methods}.
\newblock \bibinfo{journal}{\emph{Journal of Machine Learning Research}}
  \bibinfo{volume}{8}, \bibinfo{number}{May} (\bibinfo{year}{2007}),
  \bibinfo{pages}{1007--1025}.
\newblock


\bibitem[\protect\citeauthoryear{Wainberg, Merico, Delong, and Frey}{Wainberg
  et~al\mbox{.}}{2018}]%
        {WMD18DLbiomedicine}
\bibfield{author}{\bibinfo{person}{Michael Wainberg}, \bibinfo{person}{Daniele
  Merico}, \bibinfo{person}{Andrew Delong}, {and} \bibinfo{person}{Brendan~J
  Frey}.} \bibinfo{year}{2018}\natexlab{}.
\newblock \showarticletitle{Deep learning in biomedicine}.
\newblock \bibinfo{journal}{\emph{Nature Biotechnology}} \bibinfo{volume}{36},
  \bibinfo{number}{9} (\bibinfo{year}{2018}), \bibinfo{pages}{829--838}.
\newblock
\showISBNx{1546-1696}
\urldef\tempurl%
\url{https://doi.org/10.1038/nbt.4233}
\showDOI{\tempurl}


\bibitem[\protect\citeauthoryear{Wang, Xu, Xu, Ma, and Lu}{Wang
  et~al\mbox{.}}{2020}]%
        {wang2020DISSECTOR}
\bibfield{author}{\bibinfo{person}{Huiyan Wang}, \bibinfo{person}{Jingwei Xu},
  \bibinfo{person}{Chang Xu}, \bibinfo{person}{Xiaoxing Ma}, {and}
  \bibinfo{person}{Jian Lu}.} \bibinfo{year}{2020}\natexlab{}.
\newblock \showarticletitle{DISSECTOR: Input Validation for Deep Learning
  Applications by Crossing-layer Dissection}. In
  \bibinfo{booktitle}{\emph{Proceedings of the 42st International Conference on
  Software Engineering}}. IEEE Press, \bibinfo{publisher}{ICSE'20}, 12.
\newblock


\bibitem[\protect\citeauthoryear{Wang et~al\mbox{.}}{Wang
  et~al\mbox{.}}{2019b}]%
        {transferlearning.xyz}
\bibfield{author}{\bibinfo{person}{Jindong Wang} {et~al\mbox{.}}}
  \bibinfo{year}{2019}\natexlab{b}.
\newblock \bibinfo{title}{Everything about Transfer Learning and Domain
  Adapation}.
\newblock \bibinfo{howpublished}{\url{http://transferlearning.xyz}}.
\newblock


\bibitem[\protect\citeauthoryear{Wang, Dong, Sun, Wang, and Zhang}{Wang
  et~al\mbox{.}}{2019a}]%
        {wang2019adversarial}
\bibfield{author}{\bibinfo{person}{Jingyi Wang}, \bibinfo{person}{Guoliang
  Dong}, \bibinfo{person}{Jun Sun}, \bibinfo{person}{Xinyu Wang}, {and}
  \bibinfo{person}{Peixin Zhang}.} \bibinfo{year}{2019}\natexlab{a}.
\newblock \showarticletitle{Adversarial sample detection for deep neural
  network through model mutation testing}. In
  \bibinfo{booktitle}{\emph{Proceedings of the 41st International Conference on
  Software Engineering}}. IEEE Press, \bibinfo{publisher}{ICSE'19},
  \bibinfo{pages}{1245--1256}.
\newblock


\bibitem[\protect\citeauthoryear{Wang, Huang, and Schneider}{Wang
  et~al\mbox{.}}{2014}]%
        {wang14icml}
\bibfield{author}{\bibinfo{person}{Xuezhi Wang}, \bibinfo{person}{Tzu-Kuo
  Huang}, {and} \bibinfo{person}{Jeff Schneider}.}
  \bibinfo{year}{2014}\natexlab{}.
\newblock \showarticletitle{Active Transfer Learning under Model Shift}. In
  \bibinfo{booktitle}{\emph{Proceedings of the 31st International Conference on
  Machine Learning}} \emph{(\bibinfo{series}{Proceedings of Machine Learning
  Research}, \bibinfo{number}{2})}, \bibfield{editor}{\bibinfo{person}{Eric~P.
  Xing} {and} \bibinfo{person}{Tony Jebara}} (Eds.). \bibinfo{publisher}{PMLR},
  \bibinfo{address}{Bejing, China}, \bibinfo{pages}{1305--1313}.
\newblock
\urldef\tempurl%
\url{http://proceedings.mlr.press/v32/wangi14.html}
\showURL{%
\tempurl}


\bibitem[\protect\citeauthoryear{Zadrozny and Elkan}{Zadrozny and
  Elkan}{2001}]%
        {zadrozny2001obtaining}
\bibfield{author}{\bibinfo{person}{Bianca Zadrozny} {and}
  \bibinfo{person}{Charles Elkan}.} \bibinfo{year}{2001}\natexlab{}.
\newblock \showarticletitle{Obtaining calibrated probability estimates from
  decision trees and naive Bayesian classifiers}. Citeseer,
  \bibinfo{publisher}{Citeseer}, 7.
\newblock


\bibitem[\protect\citeauthoryear{Zadrozny and Elkan}{Zadrozny and
  Elkan}{2002}]%
        {zadrozny2002transforming}
\bibfield{author}{\bibinfo{person}{Bianca Zadrozny} {and}
  \bibinfo{person}{Charles Elkan}.} \bibinfo{year}{2002}\natexlab{}.
\newblock \showarticletitle{Transforming classifier scores into accurate
  multiclass probability estimates}. In \bibinfo{booktitle}{\emph{Proceedings
  of the eighth ACM SIGKDD international conference on Knowledge discovery and
  data mining}}. ACM, \bibinfo{publisher}{KDD}, \bibinfo{pages}{694--699}.
\newblock


\bibitem[\protect\citeauthoryear{Zech, Badgeley, Liu, Costa, Titano, and
  Oermann}{Zech et~al\mbox{.}}{2018}]%
        {Zech18VarGenDLRadiograph}
\bibfield{author}{\bibinfo{person}{John~R. Zech}, \bibinfo{person}{Marcus~A.
  Badgeley}, \bibinfo{person}{Manway Liu}, \bibinfo{person}{Anthony~B. Costa},
  \bibinfo{person}{Joseph~J. Titano}, {and} \bibinfo{person}{Eric~Karl
  Oermann}.} \bibinfo{year}{2018}\natexlab{}.
\newblock \showarticletitle{Variable generalization performance of a deep
  learning model to detect pneumonia in chest radiographs: A cross-sectional
  study}.
\newblock \bibinfo{journal}{\emph{PLOS Medicine}} \bibinfo{volume}{15},
  \bibinfo{number}{11} (\bibinfo{date}{11} \bibinfo{year}{2018}),
  \bibinfo{pages}{1--17}.
\newblock
\urldef\tempurl%
\url{https://doi.org/10.1371/journal.pmed.1002683}
\showDOI{\tempurl}


\bibitem[\protect\citeauthoryear{Zhang, Harman, Ma, and Liu}{Zhang
  et~al\mbox{.}}{2019}]%
        {zhang2019surveyMLT}
\bibfield{author}{\bibinfo{person}{Jie~M. Zhang}, \bibinfo{person}{Mark
  Harman}, \bibinfo{person}{Lei Ma}, {and} \bibinfo{person}{Yang Liu}.}
  \bibinfo{year}{2019}\natexlab{}.
\newblock \showarticletitle{Machine Learning Testing: Survey, Landscapes and
  Horizons}.
\newblock \bibinfo{journal}{\emph{CoRR}}  \bibinfo{volume}{abs/1906.10742}
  (\bibinfo{year}{2019}), 35.
\newblock
\showeprint[arxiv]{1906.10742}
\urldef\tempurl%
\url{http://arxiv.org/abs/1906.10742}
\showURL{%
\tempurl}


\bibitem[\protect\citeauthoryear{Zhang, Zhang, Zhang, Liu, and Khurshid}{Zhang
  et~al\mbox{.}}{2018}]%
        {zhang2018deeproad}
\bibfield{author}{\bibinfo{person}{Mengshi Zhang}, \bibinfo{person}{Yuqun
  Zhang}, \bibinfo{person}{Lingming Zhang}, \bibinfo{person}{Cong Liu}, {and}
  \bibinfo{person}{Sarfraz Khurshid}.} \bibinfo{year}{2018}\natexlab{}.
\newblock \showarticletitle{DeepRoad: GAN-based Metamorphic Testing and Input
  Validation Framework for Autonomous Driving Systems}. In
  \bibinfo{booktitle}{\emph{Proceedings of the 33rd ACM/IEEE International
  Conference on Automated Software Engineering}} (Montpellier, France)
  \emph{(\bibinfo{series}{ASE 2018})}. \bibinfo{publisher}{ACM},
  \bibinfo{address}{New York, NY, USA}, \bibinfo{pages}{132--142}.
\newblock
\showISBNx{978-1-4503-5937-5}
\urldef\tempurl%
\url{https://doi.org/10.1145/3238147.3238187}
\showDOI{\tempurl}


\bibitem[\protect\citeauthoryear{Zhou}{Zhou}{2016}]%
        {zhou2016learnware}
\bibfield{author}{\bibinfo{person}{Zhi-Hua Zhou}.}
  \bibinfo{year}{2016}\natexlab{}.
\newblock \showarticletitle{Learnware: On the Future of Machine Learning}.
\newblock \bibinfo{journal}{\emph{Front. Comput. Sci.}} \bibinfo{volume}{10},
  \bibinfo{number}{4} (\bibinfo{date}{Aug.} \bibinfo{year}{2016}),
  \bibinfo{pages}{589--590}.
\newblock
\showISSN{2095-2228}
\urldef\tempurl%
\url{https://doi.org/10.1007/s11704-016-6906-3}
\showDOI{\tempurl}


\bibitem[\protect\citeauthoryear{Zhou}{Zhou}{2019}]%
        {zhou2019abductive}
\bibfield{author}{\bibinfo{person}{Zhi-Hua Zhou}.}
  \bibinfo{year}{2019}\natexlab{}.
\newblock \showarticletitle{Abductive learning: Towards bridging machine
  learning and logical reasoning}.
\newblock \bibinfo{journal}{\emph{Science China Information Sciences}}
  \bibinfo{volume}{62}, \bibinfo{number}{7} (\bibinfo{year}{2019}),
  \bibinfo{pages}{76101}.
\newblock


\bibitem[\protect\citeauthoryear{Zhu, Wang, Tsou, and Ma}{Zhu
  et~al\mbox{.}}{2009}]%
        {zhu2009active}
\bibfield{author}{\bibinfo{person}{Jingbo Zhu}, \bibinfo{person}{Huizhen Wang},
  \bibinfo{person}{Benjamin~K Tsou}, {and} \bibinfo{person}{Matthew Ma}.}
  \bibinfo{year}{2009}\natexlab{}.
\newblock \showarticletitle{Active learning with sampling by uncertainty and
  density for data annotations}.
\newblock \bibinfo{journal}{\emph{IEEE Transactions on audio, speech, and
  language processing}} \bibinfo{volume}{18}, \bibinfo{number}{6}
  (\bibinfo{year}{2009}), \bibinfo{pages}{1323--1331}.
\newblock


\bibitem[\protect\citeauthoryear{Zhu}{Zhu}{2005}]%
        {zhu2005semi}
\bibfield{author}{\bibinfo{person}{Xiaojin~Jerry Zhu}.}
  \bibinfo{year}{2005}\natexlab{}.
\newblock \bibinfo{booktitle}{\emph{Semi-supervised learning literature
  survey}}.
\newblock \bibinfo{type}{{T}echnical {R}eport}.
  \bibinfo{institution}{University of Wisconsin-Madison Department of Computer
  Sciences}.
\newblock


\end{thebibliography}

\end{document}